\pdfoutput=1
\documentclass[journal]{IEEEtran}

\usepackage{graphicx}
\usepackage{siunitx}
\usepackage{url}
\graphicspath{{figures/}}

\begin{document}
\title{Can Language Models See the Circuit in Pixelated RFIC Passives?} 

\author{Jonathan~Zhou,~\IEEEmembership{Graduate Student Member,~IEEE,}
        Kaushik~Sengupta,~\IEEEmembership{Fellow,~IEEE}%
\thanks{J. Zhou and K. Sengupta are with the Department of Electrical and Computer Engineering, Princeton University,
Princeton, NJ 08544 USA.}}

\markboth{Preprint}%
{Zhou and Sengupta: Can Language Models See the Circuit in Pixelated RFIC Passives?}

\maketitle

\begin{abstract}
The emergence of inverse-designed, pixelated passives for microwave and millimeter-wave applications has been
a major development in the field of RFIC design. These devices, which appear to be QR-code-like in structure, 
exist in a generalizable, high-dimensional design space that includes conventional passive topologies. However, these 
structures are naturally unintuitive for designers, and failures or deviations in performance are difficult to 
debug. In this work, we ask whether a frontier language model can understand, predict, and edit the responses
and layout structure of pixelated passives. The agent, run with all tools disabled, is first asked 
to read layouts on 18$\times$18 and 27$\times$27 tile grids and, in a single reply, predict the input impedance and loss of the structure from 20-100\,GHz. On eight
GA-synthesized layouts and five random fills, the predicted Smith-chart locus follows full-wave simulation results 
closely, with a systematic shift that places features about 16\% lower in frequency than EM. On multipath loop structures it identifies notches successfully but shows a smaller frequency bias, about half that of an ideal transmission-line model that ignores bends. The agent is then asked to debug seven designs that fail their specification, and given a budget of five EM simulations and at most twelve changed tiles, it is able to improve all seven and bring five designs within specification. These experiments suggest language models can serve as an interpretability and debugging tool for traditionally non-intuitive inverse-designed passives. 

\end{abstract}

\begin{IEEEkeywords}
Large language models, inverse design, pixelated passives, millimeter-wave integrated circuits, interpretability, electromagnetic simulation.
\end{IEEEkeywords}

\section{Introduction}
\IEEEPARstart{T}{he} design of passive electromagnetic (EM) structures in radio-frequency integrated circuit (RFIC) design is critical for performance, defining bandwidth limits, losses, and amplifier modes of operation. Traditionally, the design of EM passives has relied on fixed templates and topologies which are derived from textbook theory; each structure corresponds to a single fixed function. These are constructed from lumped inductors, capacitors, and resistors, and from transmission lines. What this means is that designer knowledge and intuition is crucial, and implementing passives with new functionalities is unintuitive and difficult. 

A recent area of research seeking to address this problem is inverse design of passive structures. Inspired by inverse design in photonics~\cite{lu2013nanophotonic,piggott2015inverse,molesky2018inverse}, this domain instead treats passive structures from a top-down approach; the designer specifies desired functionality, and an optimization process synthesizes a ``pixelated'' structure. This space consists of an area divided into a grid of tiles which can be filled with metal or left blank. In this way, an optimization algorithm, often a genetic algorithm (GA), places pixels until the specification is met. Numerous works have demonstrated this approach for matching networks, combiners, dividers, and filters~\cite{karahan2023mmwave,karahan2024natcomm,guo2025dallem}, often co-designed with active circuits such as power amplifiers~\cite{karahan2023mmwave,karahan2024natcomm}. 

However, while these structures are known to work, how they work has received little study; recent near-field measurements of a pixelated filter, for example, show current patterns resembling coupled lines and stubs~\cite{zhou2026pixelfilter}. They are hard for human designers to interpret and, as a result, hard to debug. If the optimization fails or performance shifts after fabrication, there is no established method to understand what went wrong and how to fix it. 

\begin{figure*}[!t]
  \centering
  \includegraphics[width=\textwidth]{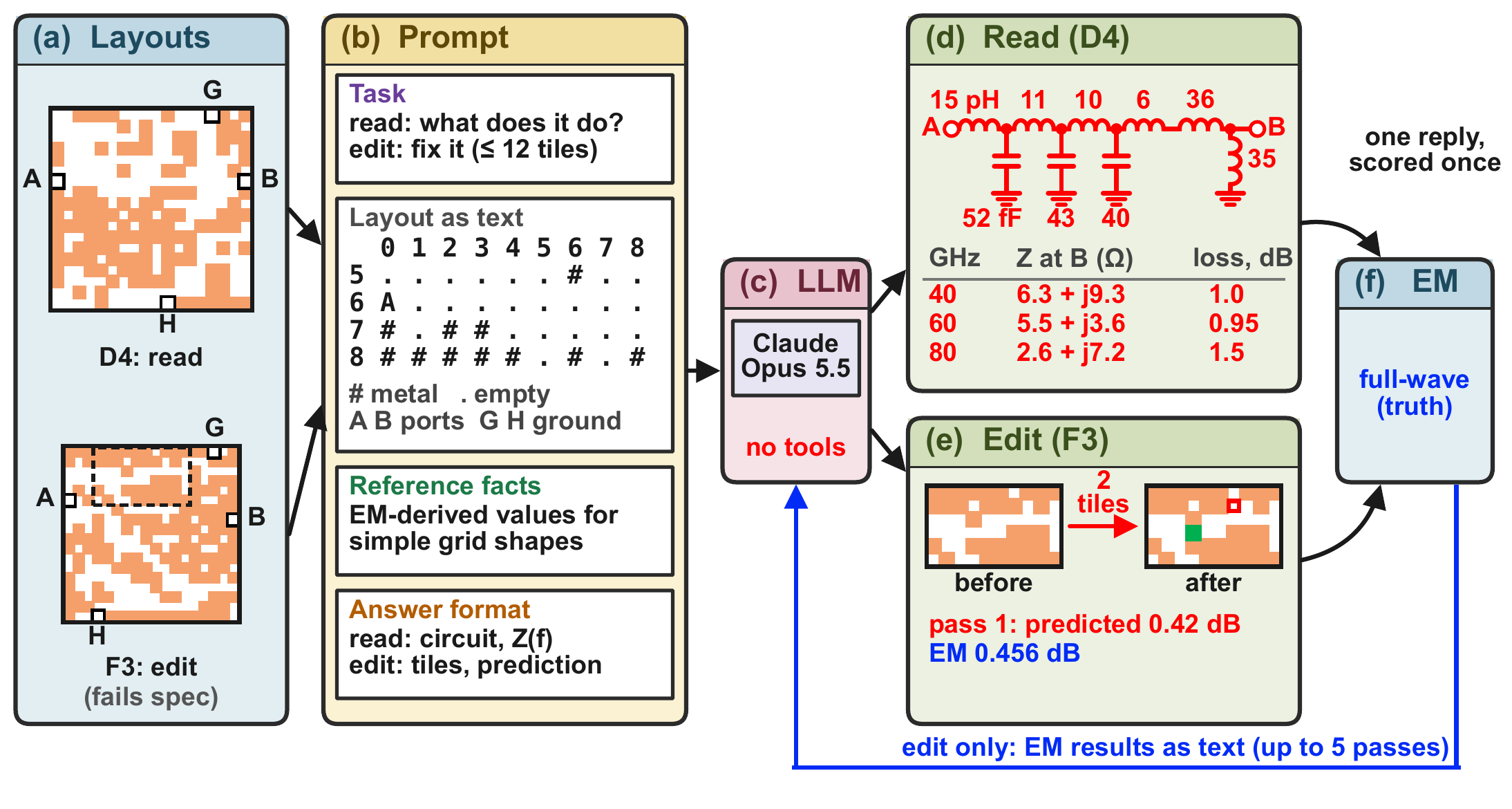}
  \caption{Overall experimental flow seeking to determine if LLMs can interpret and edit pixelated structures. (a) The pixelated structure consists of a metal grid. Details of the structures tested are found in Table\,\ref{tab:designs}. (b) Prompt construction consists primarily of a task (read or edit), the layout represented in text form, reference facts for simple shapes in this stackup, and the required answer from the agent. (c) The LLM used in this study is Claude Opus 5.5. (d) In read cases the answer is an equivalent circuit and predicted impedances and loss by frequency. (e) In edit cases that is a proposed new geometry and predicted performance. (f) An EM extraction is performed on all structures to verify and score the agent results.}
  \label{fig:Overview}
\end{figure*}
\begin{table*}[!t]
  \caption{Design sets}
  \label{tab:designs}
  \centering
  \renewcommand{\arraystretch}{1.15}
  \begin{tabular}{l c c p{0.15\textwidth} p{0.22\textwidth} p{0.27\textwidth}}
    \hline
    ID & Count & Canvas (tiles) & Band & Origin & Used for \\
    \hline
    D1--D8 & 8 & $18\times18$ & 40\,GHz (D1--D5), 80\,GHz (D6--D8) & GA matching networks, 40--54\% fill & Reading, one reply (Figs.~\ref{fig:interpretation}, \ref{fig:prediction}) \\
    R1--R5 & 5 & $18\times18$ & 20--100\,GHz & Random fill, 34--62\% & Reading control: no designed structure (Fig.~\ref{fig:prediction}) \\
    S0 & 1 & $18\times18$ & 20--100\,GHz & Straight one-tile line & Reading control: known answer (Fig.~\ref{fig:prediction}) \\
    L1--L8 & 8 & $27\times27$ & 10--110\,GHz & Hand-drawn loops and parallel paths & Reading of multipath notches (Figs.~\ref{fig:interpretation}, \ref{fig:loops}) \\
    F1--F7 & 7 & $18\times18$ & 40\,GHz (F1--F4), 80\,GHz (F5--F7) & GA designs that fail their specification & Debugging, $\le$5 EM passes (Fig.~\ref{fig:debug}) \\
    \hline
  \end{tabular}
\end{table*}

In parallel, large language models (LLM) have become ubiquitous in the last few years as a tool to assist in coding~\cite{chen2021codex,jimenez2024swebench}. In the space of integrated circuit (IC) design, works have explored using LLMs for digital design and RTL generation~\cite{liu2023verilogeval,lu2024rtllm,liu2023chipnemo}, analog design~\cite{lai2025analogcoder,yin2024adollm,chen2025menter}, and RFIC design~\cite{rfagent2026,rfampdesigner2026,tacsrf2026}. These RFIC efforts range from RF question-answering datasets for fine-tuning and retrieval~\cite{rfagent2026} to agents that size fixed netlist topologies~\cite{rfampdesigner2026} or drive commercial CAD tools for simulation, EM, layout, and extraction~\cite{tacsrf2026}. Where passives are designed, the model tunes the values or dimensions of standard templates used by human designers rather than drawing the structure itself. There has been some work applying LLMs to two-dimensional EM structures~\cite{zhang2025chat2chip}. There, a model is fine-tuned on simulated designs and maps freeform metasurfaces, encoded as text by a grid of control points, to a transmission spectrum and back. Concurrently, RFBench~\cite{cui2026rfbench} introduced a benchmark of procedurally generated pixelized PCB passives (1--10\,GHz), each a single feed-to-feed connected trace with no floating metal and no corner-only contacts, on which multimodal models, with and without fine-tuning, predict S-parameters directly and propose one-shot redesigns. Inverse-designed layouts, by contrast, contain floating islands and tiles that touch only at corners, which the reader must resolve. Our study asks something different of the model: it must read an inverse-designed mm-wave IC passive as an equivalent circuit. That circuit is an explanation a designer can check and reason with, not only a route to a number, and it lets us separate errors in reading the layout from errors in calculation. The agent then uses this reading to repair failing designs with small edits under EM feedback. To our knowledge, no prior work has examined whether a language model can read inverse-designed passives as equivalent circuits and use that reading to repair failing designs with small, EM-verified edits.

Inspired by recent work on weight-space programming~\cite{takehi2026wsp}, our work posits that if frontier LLMs can read neural-network weights and write weights that implement a specified behavior, often without tools, by first reasoning about the program a network should implement and then converting it into weights~\cite{goworek2026replicating}, it is possible that they can understand pixelated EM structures through an analogous intermediate representation: an equivalent circuit.
We present a frontier LLM with a series of tasks designed to test if it can do the following:

\begin{enumerate}
    \item Read a text-representation of a pixelated structure and understand its signal paths, port connections, and connectivity. 
    \item Build an accurate, equivalent circuit of the pixelated structure. This includes mapping each pixel to a specific device and functionality. 
    \item Predict port impedances or resonant notch locations as a function of frequency.
    \item Utilize its equivalent circuit understanding to identify what causes faulty structures to miss specification.
    \item Propose judicious edits to improve performance of pixelated structures in cases they miss specification.
\end{enumerate}

\begin{figure*}[!t]
  \centering
  \includegraphics[width=\textwidth]{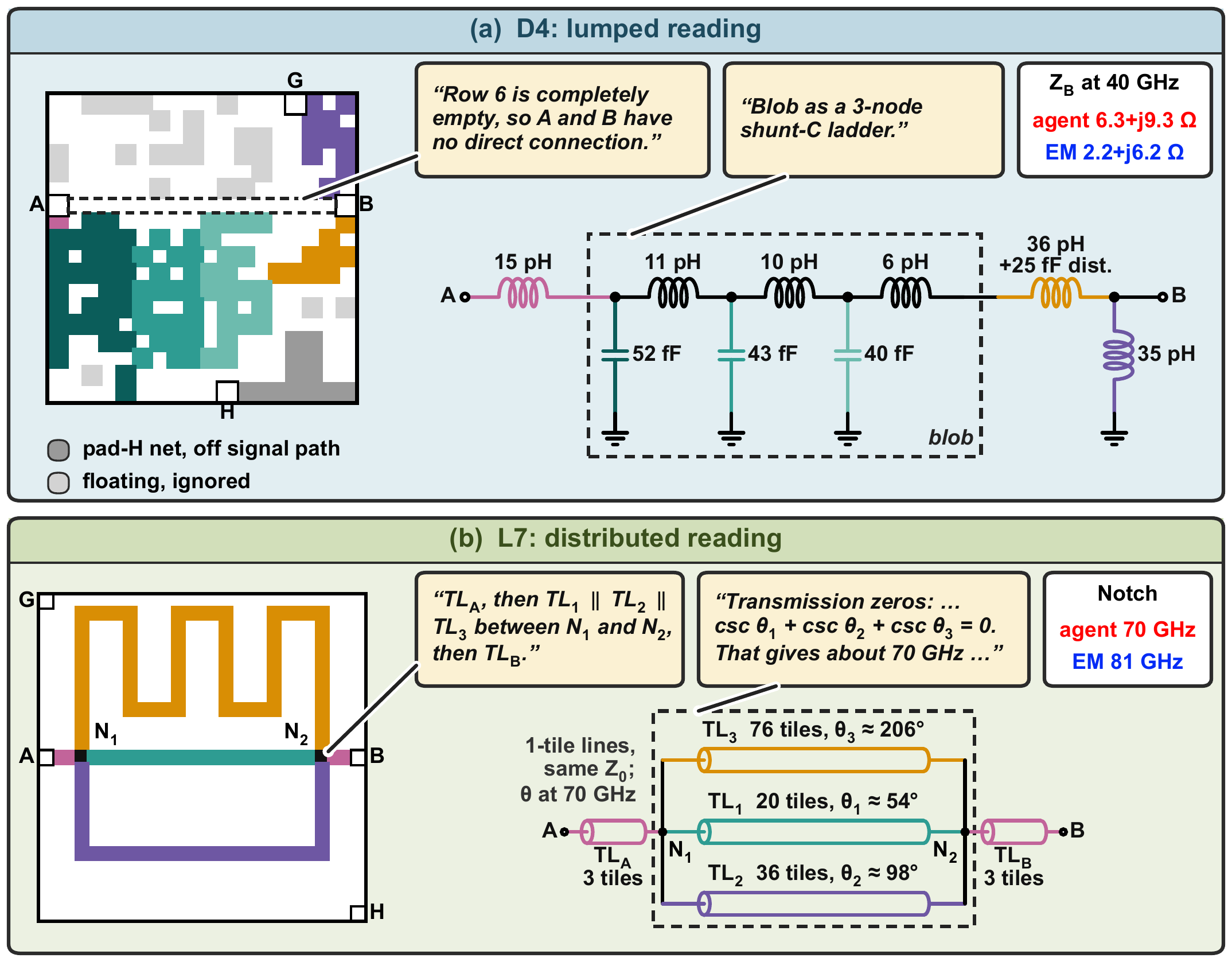}
  \caption{Structure interpretation and circuit modeling. (a) D4 is interpreted by the agent as a lumped element network, which chooses to group pixels by function. Tile coloring is transcribed from the agent's text. Grey pixels were 
  identified as floating or as ground-pad net and  discarded by the agent. The white box shows the agent's predicted Z at 40\,GHz versus EM simulation.
  (b) L7 is interpreted as a distributed network: three transmission lines in parallel between nodes N$_1$ and N$_2$. The agent predicts the notch where the signals through the three paths cancel at N$_2$, i.e.\ where $\sum_i \csc\theta_i = 0$ for the electrical lengths $\theta_i$ of the three lines (equal $Z_0$), giving 70\,GHz. Snippets of the agent's reasoning are shown verbatim in the text boxes. 
  The white box shows the agent's predicted notch frequency versus EM simulation.}
  \label{fig:interpretation}
\end{figure*}
We find that the agent reads the connectivity of every layout correctly (22/22) and predicts impedance contours that follow EM in shape: the mean $|\Delta\Gamma|$ is 0.30 on the GA designs, against 0.76 for a \SI{50}{\ohm} guess. Its predicted features sit about 16\% low in frequency, and it misjudges the design frequency of all five 40\,GHz designs. On multipath loops it finds every EM notch within $\pm$15\%. Given five EM simulations and at most twelve changed tiles, it improves all seven failing designs and brings five within specification. Section~\ref{sec:read} covers reading, Section~\ref{sec:debug} editing and debugging, and Section~\ref{sec:discussion} cost and limitations.

\section{Pixel Layout Interpretation and Prediction}
\label{sec:read}

\begin{figure*}[!t]
  \centering
  \includegraphics[width=\textwidth]{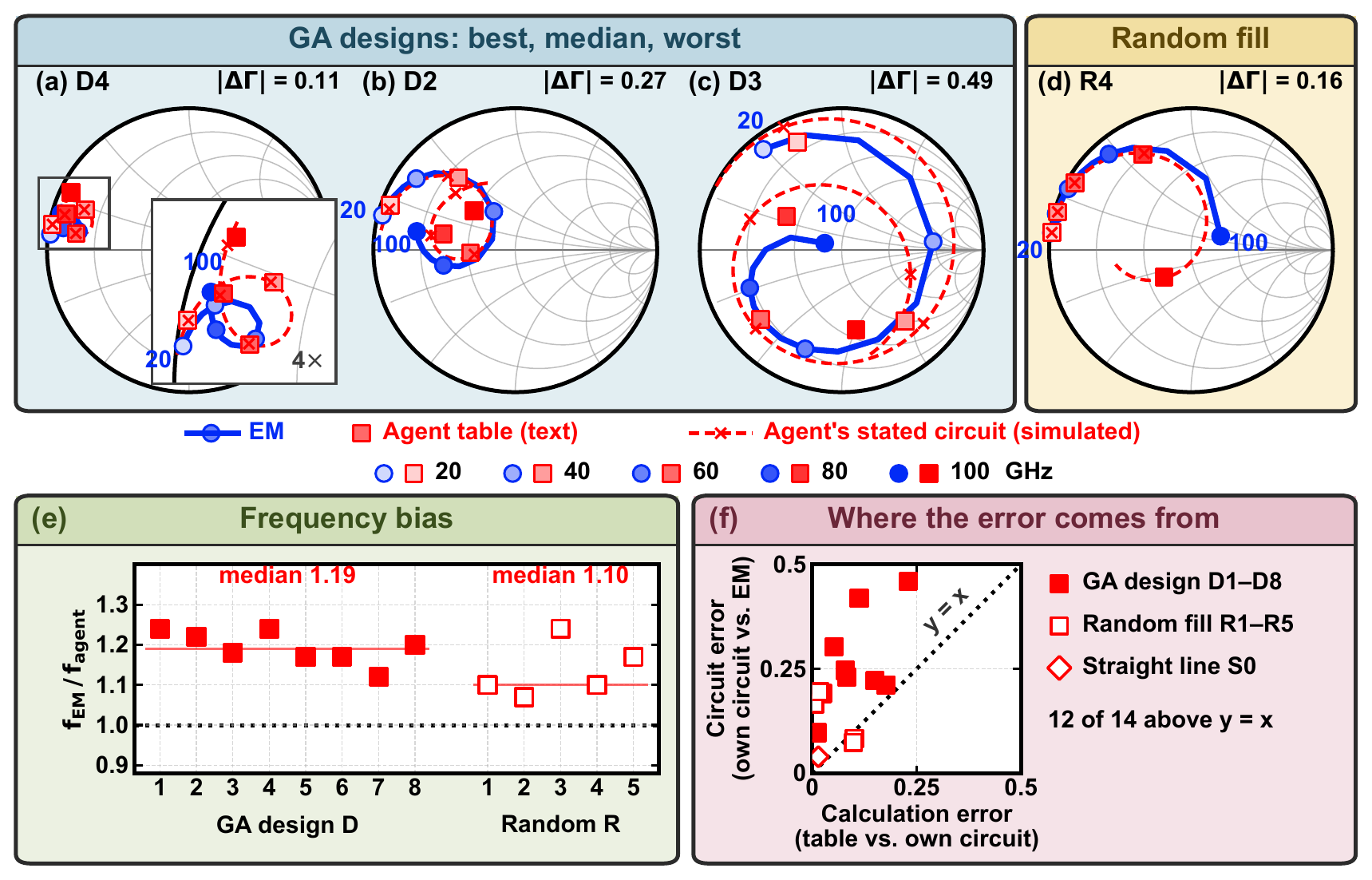}
  \caption{Impedance contour prediction. (a)--(c) Impedance into port B (port A terminated in \SI{50}{\ohm}), 20--100\,GHz,
  for the GA designs with the best, median, and worst mean prediction error $|\Delta\Gamma|$ (Smith-chart distance, top right).
  Blue: EM. Red squares: the agent's predictions, as written in its answer. Red dashed: our simulation of the circuit stated in
  the agent's answer (values verbatim, nothing tuned); red $\times$: that circuit at the five
  predicted frequencies. Marker shade darkens with frequency. (d) The same for a random-fill layout (R4).
  (e) Frequency scale $s^* = f_\mathrm{EM}/f_\mathrm{agent}$ that best maps each prediction onto the EM contour; lines: group medians.
  (f) Calculation error (table vs.\ own circuit) against circuit error (circuit vs.\ EM), both mean $|\Delta\Gamma|$;
  dotted: $y = x$. Points above the line: the error comes mainly from the circuit, not the calculation.}
  \label{fig:prediction}
\end{figure*}
\subsection{Prompt Construction}
\label{subsec:prompt}
For all experiments, the model used was Claude Opus 5.5 (\texttt{claude-opus-5-5}) at ``xhigh'' reasoning effort, run through the Claude Code command-line interface in non-interactive mode with all tools disabled, no MCP servers or user settings loaded, and the default system prompt replaced by ``Answer the user's request.'' Each run started from an empty working directory. No code, calculator, simulator, or memory calls were permitted. Each read task/layout instructed the model to provide only one reply at the end, while debug tasks had a budget of replies with EM feedback. All runs were audited to confirm that no tools were ever invoked. All EM results, both those given in prompts and those used for scoring, come from the planar full-wave solver EMX. 

Each task is constructed as a single text prompt with seven common parts, summarized in the following structure:

\begin{enumerate}
    \item \textbf{Task.} Infer what the passive structure does from its layout alone (read) or debug an existing passive structure from its layout alone (edit). The prompt states that no tools are available and that closed-form transmission-line and circuit reasoning are allowed and expected.
    \item \textbf{Structure.} An N$\times$N grid of square metal tiles on a single planar layer with certain pitch and width is given. Adjacent tiles have a certain overlap so shared edges merge into a contiguous structure. The technology and metal and ground layers are provided as well. 
    \item \textbf{Ports.} Port A (P1, \SI{50}{\ohm}) and Port B (P2) are on opposite edges with a location specified by row, column format. Two grounded ports and their locations are also specified. 
    \item \textbf{Line model.} EM-simulated $Z_0$, $\epsilon_\mathrm{eff}$, and loss of plain straight lines over a range of physical widths on this stack, given in place of textbook microstrip formulas.
    \item \textbf{Primitives Table.} EM-simulated $Z_0$, $\epsilon_\mathrm{eff}$, and loss of straight grid lines 1, 2, 3, 4, and 6 tiles wide, and equivalent-circuit values for five simple grid shapes: corner neck, diagonal staircase, one-tile gap, coupled pair, and a shorted stub. These are specific to the tile grid and take precedence over the line model where the two overlap. A discussion of the function of the table is provided in Section~\ref{subsec:table}.
    \item \textbf{Origin.} A one-line note on how the layout was made. The random-fill controls received the same note as the GA designs (that a genetic algorithm placed the tiles).
    \item \textbf{Layout.} The tile grid as text, one row per line, with \texttt{\#} for metal, \texttt{.} for empty, and the letters for A, B, G, H for the RF ports and ground ports (Fig.~\ref{fig:Overview}(b)).
\end{enumerate}

The agent then answers in fixed sections. For the read task, the reply consists of observations (signal paths, ground connections, floating metal, DC behavior); an equivalent circuit representation it selected; estimated design frequency; the predicted impedance seen from port B with port A terminated in \SI{50}{\ohm}; a table of predicted impedances and transducer loss (available source power over power delivered to the load) at 20, 40, 60, 80, and 100\,GHz; and its own confidence level. For loop structure read tasks, the model also estimates the frequency of notches in $|S_{21}|$. 

For the debugging task (Section~\ref{sec:debug}), the prompt adds editing rules, a specification, and EM results of the current proposed design. In turn, the agent reports a diagnosis, a predicted loss and impedance for its edit, and the new layout in the same text format. 

\subsection{Structure Interpretation}
\begin{figure*}[!t]
  \centering
  \includegraphics[width=\textwidth]{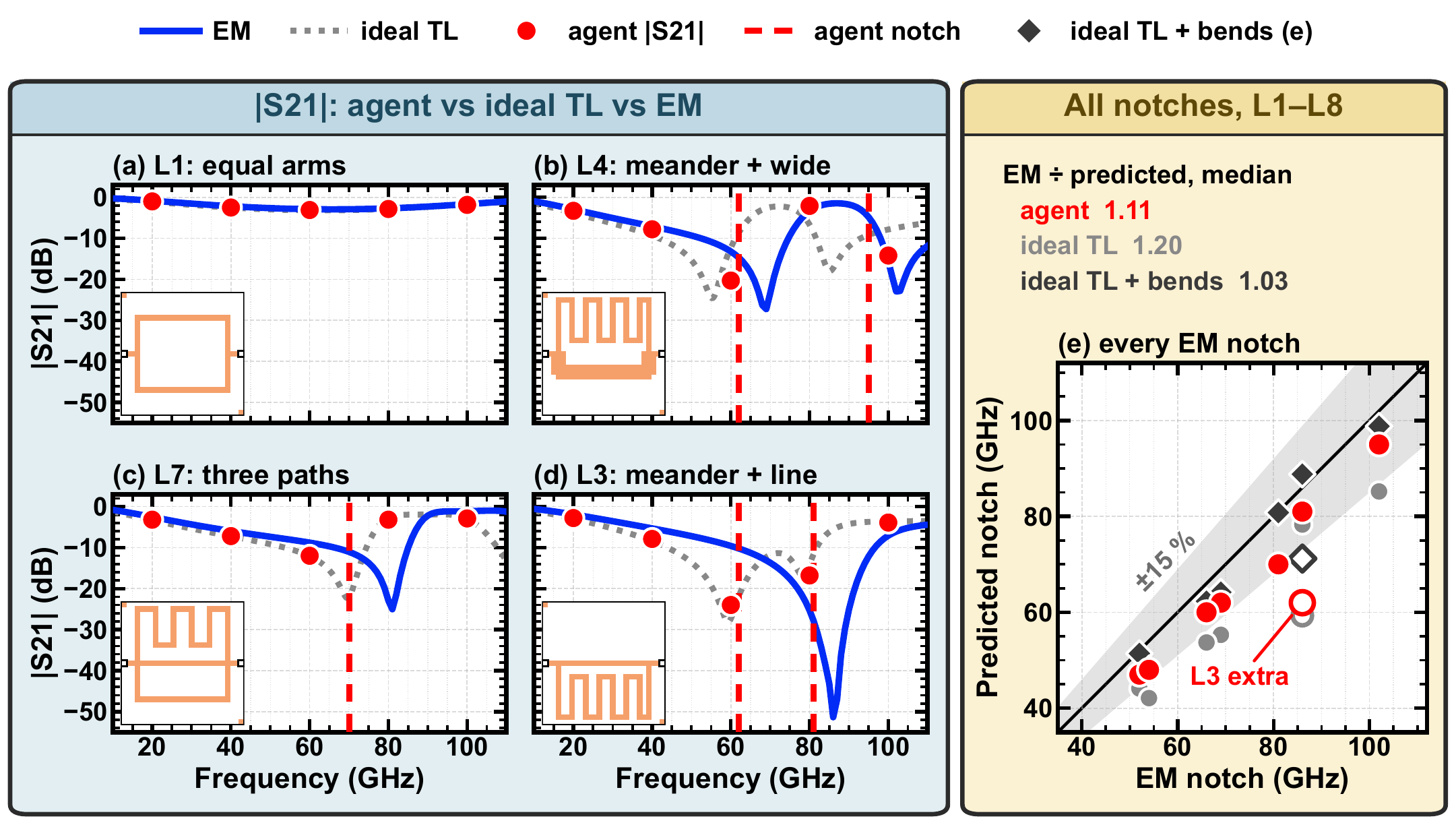}
  \caption{Loop prediction results. (a)--(d) The agent is asked to interpret a variety of multi-path transmission line loops. The agent's predictions are compared to EM simulations, and in most cases the agent tracks ideal transmission line theory more closely than the EM results. 
  (e) The agent is asked to predict the notch frequency of the loop structures. The agent places notches low (median $f_\mathrm{EM}/f_\mathrm{agent}$ 1.11), between an ideal transmission-line model of the drawn centerlines (1.20) and the same model with bend corrections (1.03).}
  \label{fig:loops}
\end{figure*}
Although pixelated passive structures allow for generalization of functionality and rapid, automated synthesis~\cite{karahan2023mmwave,karahan2024natcomm}, they remain non-intuitive for designers. These structures contain multiple paths and stubs of varying length, holes of varying sizes, and floating metals which complicate any analysis or interpretation. 

For the structure interpretation experiment, the prompt structure is described in Section~\ref{subsec:prompt}. From Table~\ref{tab:designs}, the first four rows are used to provide 22 different examples. In Fig.~\ref{fig:interpretation}(a), the example is a pixel matching network from GA optimization which presents an impedance of $2.2 + j6.2$\,\si{\ohm} at 40\,GHz. The agent first makes a series of high-level observations: there is no direct, straight-line connection from Port A to B; the metal tied to ground port H lies off the signal path and is negligible; floating metal also has no effect. It identifies it generally as \textbf{``a low-pass ladder impedance step-down followed by a shunt-L at the output.''}

At this point, the agent constructs the equivalent circuit, deciding where to separate the metal into distinct components. Interestingly, it prefers to represent the structure as a lumped RLC model rather than transmission-line based. Metal segments are interpreted as largely capacitive structures. Finally, it predicts the target frequency as 55\,GHz and gives a predicted impedance of $6.3 + j9.3$\,\si{\ohm} at 40\,GHz with 1\,dB loss. The design-frequency estimate is the weakest part of the reads: across the eight GA designs it falls within $\pm$15\% of the true value for all three 80\,GHz designs but none of the five 40\,GHz designs, which it places at 55--90\,GHz. 

Example L7 (Fig.~\ref{fig:interpretation}(b)) performs the same task on a transmission-line based loop structure, one of the multipath designs studied in Section~\ref{subsec:loops}. The agent identifies multiple side paths in addition to the main path, along with meandering and a column symmetry. It creates an equivalent circuit model of transmission lines of varying widths and lengths, choosing to 
predict the notch by estimating where the signal through the three paths cancel. In this case, the notch is estimated at 70\,GHz compared to the EM-simulated 81\,GHz. 

From these examples, it can be seen that the agent is able to correctly understand connectivity, floating metals, and ground ports. The numerical predictions of operation frequency, impedance, and notch frequency are not necessarily accurate.

\subsection{Impedance Contour Prediction}

For a more specific numerically-guided study, we examine the specific impedance contours predicted by the agent for the same examples. The selected structures are targeted for operation in the mmWave band, and thus the agent predicts the impedance seen at Port B at 20, 40, 60, 80, and 100\,GHz when Port A is terminated in \SI{50}{\ohm}. Fig.~\ref{fig:prediction}(a)--(d) shows the predicted versus EM-simulated response for three GA structures representing best, median, and worst-case prediction accuracy, and one random fill structure. In blue, the EM-simulated contour is shown with circles indicating the specific frequency points of interest. In red, the squares give the agent-predicted impedance at the frequency points of interest. The red contour specifically is created by simulating the agent's stated equivalent circuit model in its response. Values and topology are taken verbatim, no tuning is performed. It can be observed that the agent is able to predict the impedance contour shapes and locations with reasonably good accuracy even across a wide frequency band, and for very different profiles/targets. For reference, a no-skill guess of \SI{50}{\ohm} at every frequency gives a mean $|\Delta\Gamma|$ of 0.76 for the GA designs and 0.83 for the random fills, against the agent's 0.30 and 0.15 (20--100\,GHz). 

However, as shown in Fig.~\ref{fig:prediction}(e), the agent has a tendency to run ahead of the true EM response when placing frequency points. We quantify this with a frequency scale $s^*$: the factor, searched over 0.6--1.8, that minimizes the mean $|\Delta\Gamma|$ between the agent's prediction at $f$ and EM at $s^* f$. In the GA designs, it appears to have a systematic bias of approximately 1.19$\times$, with a weaker, but still present bias in the random fill cases. Rescaling frequency post-hoc by the group median for each set (1.19 for GA designs, 1.10 for random fills) reduces the mean $|\Delta\Gamma|$ over 20--80\,GHz (the points that stay inside the simulated band after rescaling) from 0.28 to 0.08 (eight GA designs) and from 0.11 to 0.06 (five random fills). This is an in-sample check, with the scale fitted and scored on the same small set of designs; this should not be extrapolated as any sort of universal bias or rule-of-thumb.

At this point, we examine the source of error in the agent predictions. Because the agent is not allowed to utilize any calculators or simulators, a potential source of error is calculation error between the stated equivalent circuit and the outputted impedance table. The other major error source is equivalent circuit error, where the agent has not correctly interpreted the layout. In Fig.~\ref{fig:prediction}(f), we compare the equivalent circuit error to the calculation error. It can be seen that the error primarily comes from circuit construction rather than from calculation.

\subsection{Effect of the Reference Table}
\label{subsec:table}
In Section~\ref{subsec:prompt}, it is noted that we provide a primitives table. This consists of $Z_0$, $\epsilon_\mathrm{eff}$, and loss characterization for straight grid lines 1, 2, 3, 4, and 6 tiles wide, plus equivalent-circuit values for five simple grid shapes: a corner neck, a diagonal staircase, a one-tile gap, a coupled pair, and a ground-shorted stub. To study the effects of providing this information, we perform an ablation study. The read tasks are repeated, with the prompts edited to remove only the primitives table. Without the table, element values were less accurate. Mean $|\Delta\Gamma|$ over the eight GA designs (20--100\,GHz) rose from 0.30 to 0.47, and the frequency shift increased from 1.19$\times$ to 1.36$\times$.
The table values were extracted from EM simulations of each shape in isolation, over a ground strip only slightly wider than the shape rather than the full ground plane of the test layouts. This overstates line loss by 1.4--2$\times$ and raises $Z_0$ and $\epsilon_\mathrm{eff}$ by a few percent. A spot check with a corrected table on three designs (two runs each) gave similar read errors (D2--D4, 20--100\,GHz: mean $|\Delta\Gamma|$ 0.24 vs.\ 0.29, median $s^*$ 1.17 vs.\ 1.22), so the conclusions here do not depend on the table error.

\subsection{Multipath Structures}
\label{subsec:loops}

It is useful to examine cases where an agent struggles to interpret circuits properly. Human designers are taught to interpret matching networks in terms of series and shunt segments in a cascade; as such, textbooks and literature focus primarily on cascaded topologies, with parallel structures being much more difficult to intuit. In the pixelated read tasks, it is observed that the agent is always able to find a primarily cascaded equivalent circuit. While holes in the pixel structure potentially suggest a multi-path network, the GA-synthesized examples do not have a significant enough hole to justify this. Thus, we design a series of looped/multi-path transmission-line structures which contain meandered lines as well. They are designed to have significant multi-path effects, which create deep notches in the $|S_{21}|$ response. 

The read tasks are repeated on these structures, and some examples are shown in Fig.~\ref{fig:loops}(a)--(d). The agent is additionally asked to name the notch frequencies. The blue curve represents the EM simulated response. In grey, an ideal transmission-line model of the same circuit is simulated (drawn centerline lengths, reference-table $Z_0$ and $\epsilon_\mathrm{eff}$), with corner effects ignored. The red circles give the agent's predicted $|S_{21}|$, and the dashed vertical lines give the predicted notch frequencies. In L1, with two equal arms effectively functioning as one thin line, the model performs well as expected and correctly identifies no significant notch. However, for designs L4 and L7, the model tends to again run fast on frequency. The predicted features shift down, and the notches sit lower than their real frequency. In example L3, the model incorrectly predicts two notches compared to only one real notch; the ideal transmission-line model shows the same two dips, so the extra notch comes from transmission-line theory rather than a misread layout. Interestingly, in most cases the agent's prediction tends close to the ideal transmission-line model (L4 is the clearest exception), giving potential insight into its methodology. 

In Fig.~\ref{fig:loops}(e), the median notch prediction error is plotted for all loop designs. The agent places notches low by a median of 1.11$\times$. The ideal transmission-line model is off by 1.20$\times$, and adding bend corrections reduces this to 1.03$\times$. The bends make each path electrically shorter than its drawn centerline; the agent partly accounts for this, but not fully. As with the pixelated read case, this bias is only characterized on a small number of samples. 

\section{Layout Editing and Debugging}
\label{sec:debug}
\begin{figure*}[!t]
  \centering
  \includegraphics[width=\textwidth]{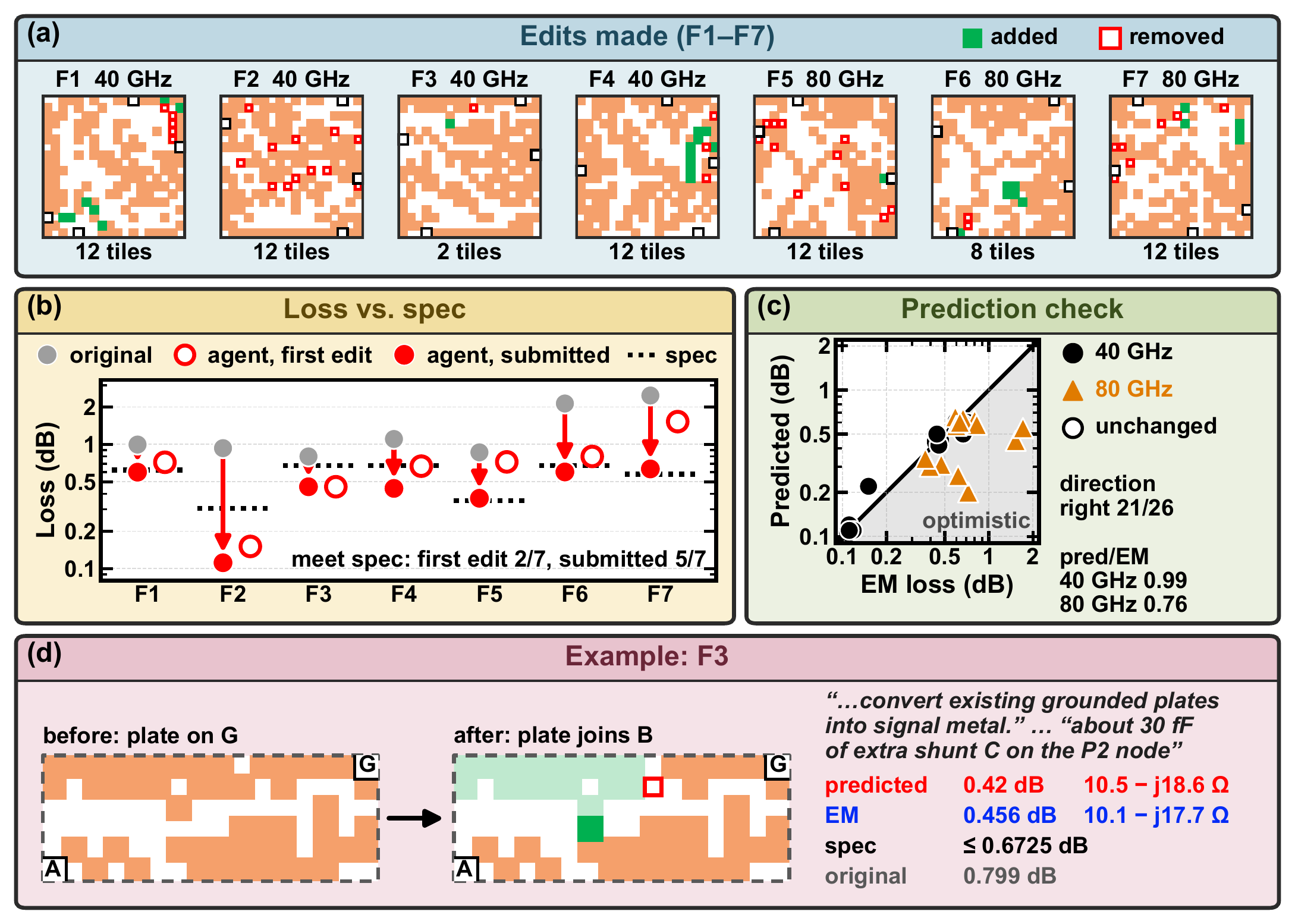}
  \caption{Debugging and editing tests. (a) The agent is presented with a failing pixel design and asked to repair the structure by adding or removing pixels to reach a specification.
  (b) Worst-case in-band loss of each original design (grey), the agent's first edit, made before any EM feedback on its own edits (hollow red), and its submitted design (filled red); dotted: loss specification. Meeting the specification also requires the return-loss limit, so F4's first edit, just under the loss line, still fails. 
  (c) The agent's predictions for structure loss are compared to the EM simulation. It can be seen 40\,GHz predictions are highly accurate, while 80\,GHz predictions are optimistic. 
  Thus it requires the EM simulation iterations for full confidence. (d) The edit of sample F3 is shown. The agent's own rationale is provided, where it identifies a chunk of metal as a 
  grounded capacitive plate, and chooses to cut it from the ground and tie it to the signal path as additional shunt capacitance.}
  \label{fig:debug}
\end{figure*}

With frontier language models demonstrating an ability to interpret and model pixelated structures, it suggests they can also perform informed debugging and editing of structures. In cases where synthesis fails to hit targets, or performance shifts during fabrication, the agent's circuit reading offers a way to understand what sections of pixelated structures do. This leads into the edit series of tasks; an agent is given a pixelated design produced by GA optimization which failed to reach a target specification. 

In addition to the prompt elements in Section~\ref{subsec:prompt}, the edit prompt is expanded to include a set of edit rules. Ports must stay where they are and at most twelve tiles may be changed from the original structure. This is to prevent the model from synthesizing/creating its own structure from scratch. Also provided is the specification: the structure must present a target impedance at port B across a narrow band (2--6\,GHz wide, centered at 40 or 80\,GHz), with worst-case in-band transducer loss and $|S_{11}|$ below set limits. The loss limits are scaled from the loss of an ideal LC match with an inductor $Q$ of 20. In these tasks, the model was not asked to provide a one-shot reply with the final design. Instead, it was allocated a budget of five replies and EM simulations. This meant the agent could propose a structure and ask for its EM simulated response as a form of feedback. As with the read task, any optimization, calculation, or other tool calls were forbidden. The agent answer format requires a diagnosis of the issue, an edited layout in text form, and the predicted loss and impedance. It is left to the agent to select the best design from its back-and-forth iteration, or to decide if a design is already satisfactory.

The test structures were selected from GA optimizations where the specifications were not met. They were chosen before any agent run, from an earlier study of failing GA designs, and share no layouts with D1--D8. The edits made to the seven test structures are shown in Fig.~\ref{fig:debug}(a). Red tiles are removed while green tiles are added. The number of edited tiles is listed below each structure, with the target design frequency above.
\begin{table*}[!t]
  \caption{What the agent can and cannot do}
  \label{tab:scorecard}
  \centering
  \renewcommand{\arraystretch}{1.15}
  \begin{tabular}{p{0.20\textwidth} p{0.27\textwidth} p{0.40\textwidth}}
    \hline
    Question & Metric & Result \\
    \hline
    Does it see the layout? & A--B path, grounded vs.\ separate pads, DC behavior, floating metal & Correct in 22/22 reads; one read missed a single one-tile floating island\\
    Does it name the design frequency? & Estimate within $\pm$15\% of the task frequency (GA) & 3/8: 80\,GHz designs 3/3, 40\,GHz designs 0/5 (placed at 55--90\,GHz)\\
    Does it predict the response? & Mean $|\Delta\Gamma|$, 20--100\,GHz (GA / random) & 0.11--0.49 / 0.09--0.21 (mean 0.30 / 0.15; \SI{50}{\ohm} guess 0.76 / 0.83) \\
    Is the error systematic? & Frequency scale $s^*$ & $>1$ on every GA and random design; median 1.19 (GA), 1.10 (random)\\
    Does its table follow its own circuit? & Calculation vs.\ circuit error, median (GA) & 0.10 vs.\ 0.24: error comes mainly from the circuit \\
    Does it read multipath structures? & Notch count; notch frequency vs.\ EM & Count right on 7/8 loops (one spurious notch, L3); all 7 EM notches found within $\pm$15\%; median $f_\mathrm{EM}/f_\mathrm{agent}$ 1.11\\
    Does it predict the effect of an edit? & Loss-change direction; predicted/EM loss & 21/26 directions right; 0.99 at 40\,GHz, 0.76 at 80\,GHz\\
    Can it fix a failing design? & Meets specification ($\le$12 tiles, 5 EM passes) & 5/7 meet, 7/7 improved; first edit alone: 2/7 meet, 7/7 improved\\
    What does the reference table add? & Read error with vs.\ without the table (GA, 20--100\,GHz) & 0.30 vs.\ 0.47 (mean); $s^*$ 1.19 vs.\ 1.36 \\
    \hline
  \end{tabular}
\end{table*}
In (b), grey dots indicate the original structure loss, with dotted black indicating the loss specification. Filled red markers give the agent's submitted design, which improves all seven structures and meets the full specification (loss and return loss) in five of seven cases. Hollow red markers give the agent's first edit, made before any EM feedback on its own edits; it already improves all seven designs and meets the specification in two. Each design was run once.

All results are EM simulated, and in (c), the agent predictions are compared to the simulation. The black dots show that 40\,GHz predictions are highly accurate, while the orange triangles indicate 80\,GHz predictions are optimistic. This indicates why the ability to verify with a small budget of EM simulations is necessary and improves performance. Finally in (d), the F3 example edits are shown, along with the model's own thinking quoted verbatim. It identifies a large portion of metal as a capacitive plate which is connected only to ground. By suggesting an edit of only two tiles, it cuts this structure from the ground port and instead connects it to the signal path. It predicts this will provide extra shunt capacitance at that node. The original structure's loss is 0.799\,dB, while the agent predicts its own edit will lower the loss to 0.42\,dB with a presented impedance of $10.5 - j18.6$\,\si{\ohm}. EM simulation supports the prediction well, giving values of 0.456\,dB loss with an impedance of $10.1 - j17.7$\,\si{\ohm}. Thus with only two judicious edits, the agent is able to substantially and accurately improve the performance.

\section{Discussion}
\label{sec:discussion}
It is apparent that frontier language models do show an ability to understand, predict, and edit pixelated passive structures. While this work does not seek to find the exact limits of agent capabilities for these tasks, we have run a series of tasks which provide meaningful insight. The qualitative question, metrics, and results are given in Table~\ref{tab:scorecard}. 

\begin{itemize}
    \item \textbf{Layout understanding:} The first base-level question is if the agent correctly identifies basic connectivity, input, output, and ground port locations, and which tiles are floating. All tasks had this question answered correctly, except one read which missed a one-tile floating island, showing%
 a basic ability to understand the structure from text representation.

    \item \textbf{Response and equivalent circuit prediction:} This question consisted of the impedance table prediction and equivalent circuit behavior. The agent showed an ability to understand pixelated structures from both GA synthesis and random fills. Additionally, our analysis showed that its proposed impedance numbers matched closely to its equivalent circuit. However, all predictions did exhibit a systematic bias in frequency of varying degrees.

    \item \textbf{Multipath structure understanding:} Structures with multiple paths were tested to see if the agent is able to reason out circuits with multiple signal paths. While the agent showed a correct understanding and identification of the paths, and was able to identify the presence of a notch correctly, it did incorrectly predict two notches in an example which only showed one real notch. Additionally, it showed a systematic frequency bias as well, and comparison to ideal transmission-line models suggests it only partly accounts for bends. 

    \item \textbf{Structure debugging ability:} In the edit tasks, 21 of 26 tests showed the agent correctly identified the direction in which loss of a structure would change based on a given edit. At 40\,GHz, the predicted loss was highly accurate, while at 80\,GHz the agent was overly optimistic. Of the seven debug tests, in all cases the agent improved the performance through a limited set of judicious changes, lowering the worst-case in-band loss by 0.87\,dB on average (median 60\%). Five satisfied the spec. In general, it seems that the agent is able to debug and edit structures successfully.
\end{itemize}

The cost of these experiments is modest. A single read took a median of 16 minutes and about 110k output tokens, or \$2.2 at API list prices; all 22 reads cost about \$50. A five-pass debug run took a median of 19 minutes of model time, plus a few minutes of EM simulation per pass, and about \$3.6; all seven cost \$33. This is small compared with the time a designer would spend analyzing a pixelated layout by hand, but slow compared with a surrogate evaluation, so the agent fits as a diagnostic and repair step after synthesis rather than inside the optimization loop.

\subsection{Limitations}
This work focused on testing general capabilities of a single model (Claude Opus 5.5) at a single effort level and analyzing where it works and where it fails. Experiments were conducted on a handful of samples for each task, with single runs for each. Repeating three read tasks twice moved the error $|\Delta\Gamma|$ of a single design by up to 0.18, so per-design numbers carry that spread. Additionally, only a single PDK technology was used for all tasks. Prompts were not designed for optimality, but rather for generality. No budget-matched baseline (for example, five EM passes of random or optimizer-driven edits under the same 12-tile limit) was run for the debug task, so the 5/7 result is not compared against a simple search of equal cost. 

A recent audit of weight-space reading, using the same model as here on small image classifiers, found that its zero-shot reading is better described as local arithmetic on the input and output layers plus priors over familiar datasets than as tracing computation through the network~\cite{ahmed2026wspaudit}; an independent replication of weight-space programming likewise found that reading weights is mostly reading the input layer~\cite{goworek2026replicating}. The analogous concerns here are whether the agent reads only parts of the layout, such as the regions near the ports, and whether it only matches familiar textbook shapes. Its stated circuits span the signal path from port to port (Figs.~\ref{fig:reader_D1}--\ref{fig:reader_L8}), and three results suggest it is not merely matching shapes. First, the agent reads random-fill layouts, which were not optimized or designed for any task, with errors of 0.09--0.21. Second, its predicted impedances follow from the equivalent circuit it builds for each layout, rather than from recalled values. Finally, removing the stackup-specific reference table raises error from 0.30 to 0.47, implying it depends on the data it is given rather than purely embedded learnings. 

Future works could look into performing more rigorous ablation studies comparing models, effort levels, and prompt levels.

\section{Conclusion}
In this work we investigated the ability of large language models to read, interpret, and debug pixelated RFIC passive structures. The agent was given a set of pixel structures which were synthesized from GA optimization, a set of randomly filled structures, and various transmission-line loops. It was asked to read the structure, understand its connectivity, propose an equivalent circuit, and to predict frequency response and impedances or notches. Across the read tasks, the agent read connectivity correctly in every case and predicted frequency responses that follow EM in shape, albeit with a systematic frequency bias. The agent was then given a series of structures which missed optimization specifications and asked to propose a small set of fixes to satisfy the targets. The agent improved all seven designs, and five met their specification. These studies indicate that frontier language models have the ability to reason about pixelated structures and decompose them into functional circuits; this in turn allows the agent to perform guided repairs and edits. This capability is a first step toward addressing a long-standing difficulty with pixel structures, and could allow designers to understand and work with them outside of a black-box approach. In conjunction with the rapid optimization and generalization allowed by these structures, such guided understanding could make inverse-designed passives practical to debug and adopt in RFIC design.

\section*{Acknowledgment}

The authors acknowledge GlobalFoundries for process design kit access, and thank Princeton ECE IMRL for support and
discussions. The authors thank Emir Ali Karahan for the pixelated inverse-design framework underlying the GA designs, and Juho Park for the EM evaluation flow and design targets used in this work. The authors are pleased to acknowledge that the work reported on in this paper was substantially performed using
Princeton University's Research Computing resources.

Claude (Anthropic) was used as a coding and writing assistant for the simulation and analysis code, the figures, and parts
of the text; it is separate from the agent evaluated in this work. The authors reviewed all content and take full
responsibility for it.

\bibliographystyle{IEEEtran}
\bibliography{references}

\appendices
\section{Supplementary Figures}
\setcounter{figure}{0}
\renewcommand{\thefigure}{S\arabic{figure}}

\begin{figure*}[!t]
  \centering
  \includegraphics[width=\textwidth]{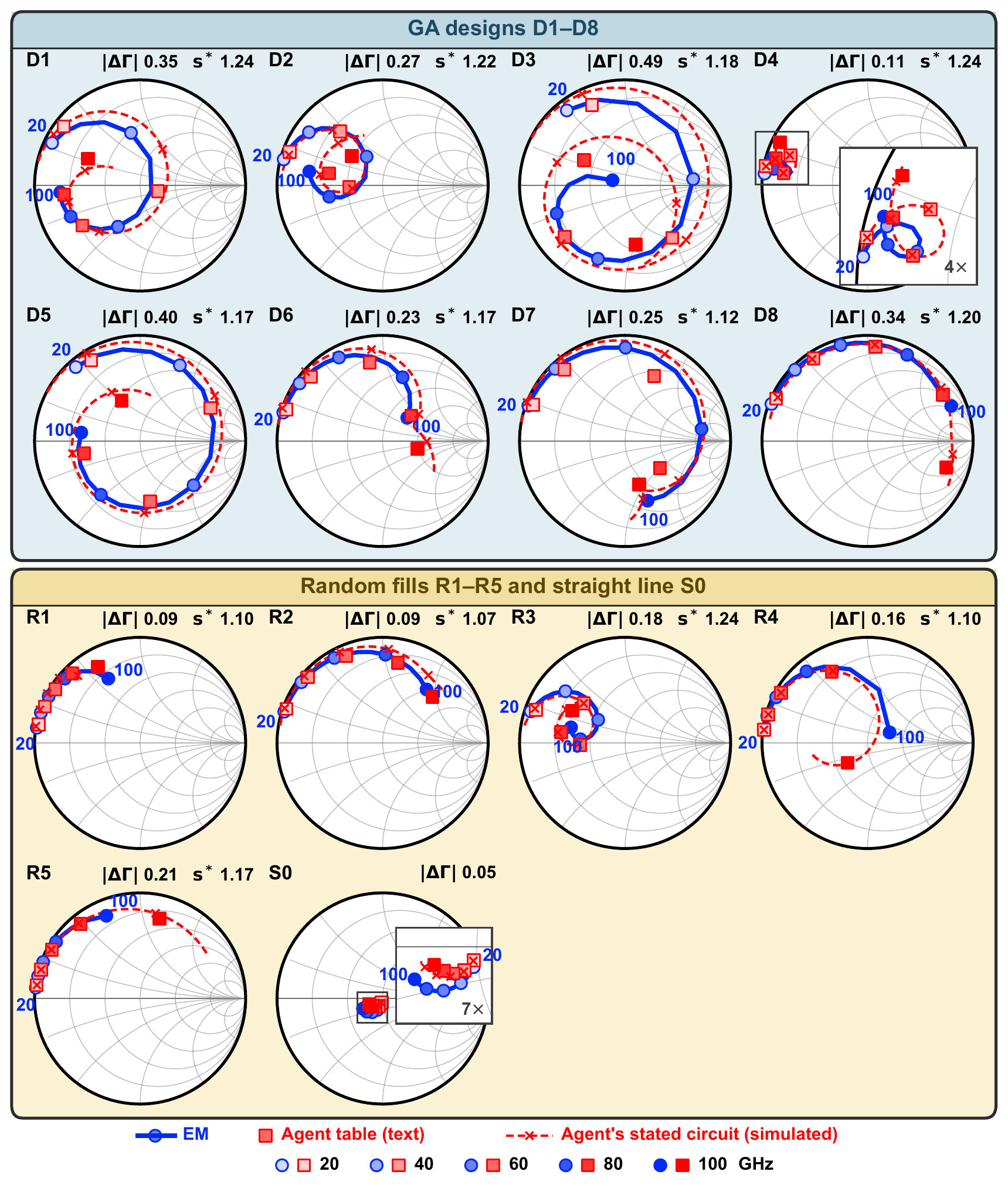}
  \caption{Impedance contour prediction for every read design: GA designs D1--D8 (top) and random fills R1--R5 plus the
  straight line S0 (bottom). Encoding as in Fig.~\ref{fig:prediction}: blue, EM; red squares, the agent's table; red dashed
  with $\times$, the agent's stated circuit simulated with its values verbatim. Each panel gives the mean $|\Delta\Gamma|$
  and the frequency scale $s^*$.}
  \label{fig:S1}
\end{figure*}

\begin{figure*}[!t]
  \centering
  \includegraphics[width=\textwidth]{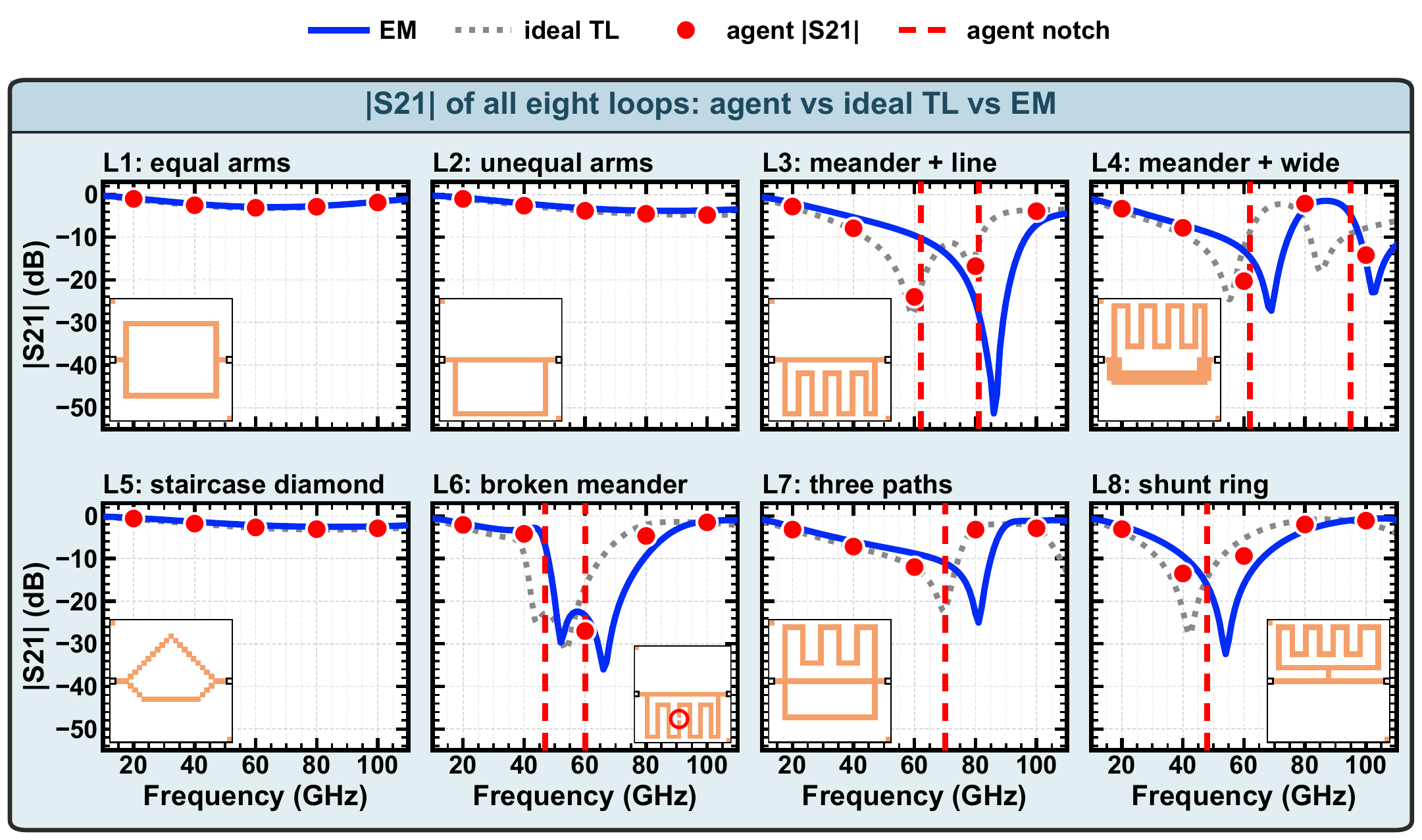}
  \caption{All eight multipath (loop) structures L1--L8. $|S_{21}|$ into \SI{50}{\ohm}: EM (blue), ideal
  transmission-line theory from the drawn centerlines (grey dotted), the agent's predicted $|S_{21}|$ at 20, 40, 60, 80
  and 100\,GHz (red dots) and its notch calls (red dashed).}
  \label{fig:S2}
\end{figure*}

\begin{figure*}[!t]
  \centering
  \includegraphics[width=\textwidth]{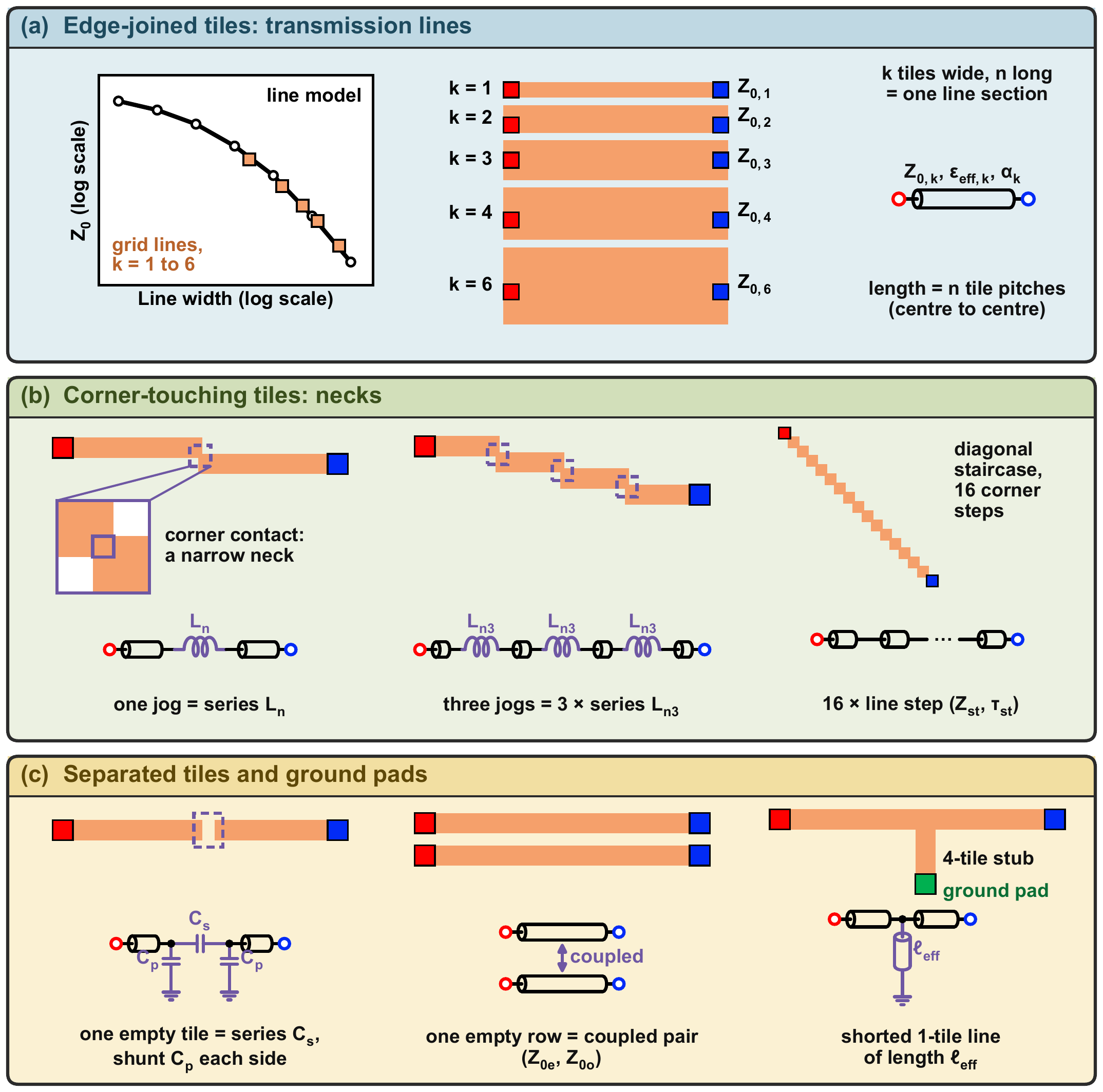}
  \caption{Grid primitives and their equivalent circuits, from full-wave EM on the pixel grid; these make up the
  reference table given to the agent. (a) Edge-joined tiles form transmission lines; a line $k$ tiles wide follows the
  generic line model. (b) Tiles touching only at a corner join through a narrow neck, a series inductance; a diagonal
  staircase acts as a line built from such steps. (c) One empty tile is a series capacitance with shunt capacitance on
  each side; two lines one empty row apart form a coupled pair; a stub to a ground pad is a shorted line.
  Port tiles red/blue, ground pad green; discontinuities in purple in both layout and circuit.}
  \label{fig:S3}
\end{figure*}

\begin{figure*}[!p]
  \centering
  \includegraphics[width=\textwidth]{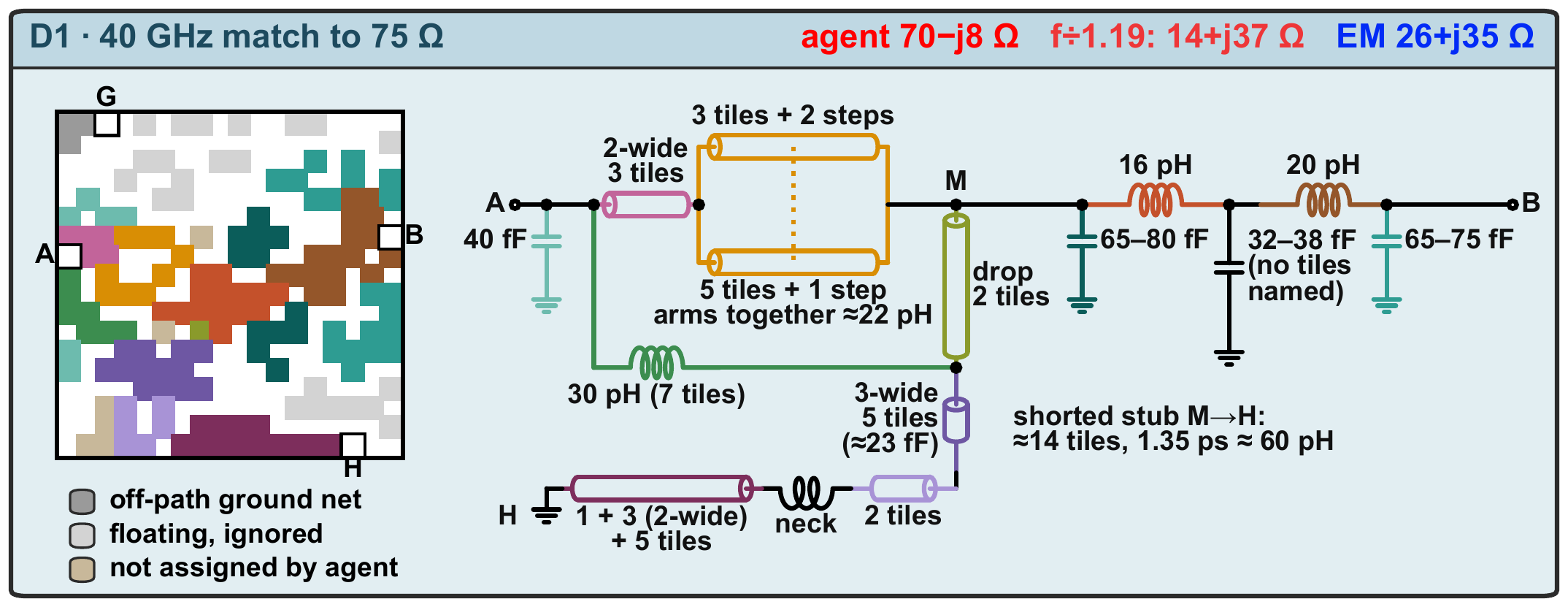}
  \caption{Agent interpretation of GA design D1. Left: layout with tiles colored by the element the agent assigned.
  Right: the agent's stated circuit. Header: $Z_\mathrm{B}$ at the task frequency from the agent's table, from its stated circuit evaluated at $f/1.19$, and from EM.}
  \label{fig:reader_D1}
\end{figure*}

\begin{figure*}[!p]
  \centering
  \includegraphics[width=\textwidth]{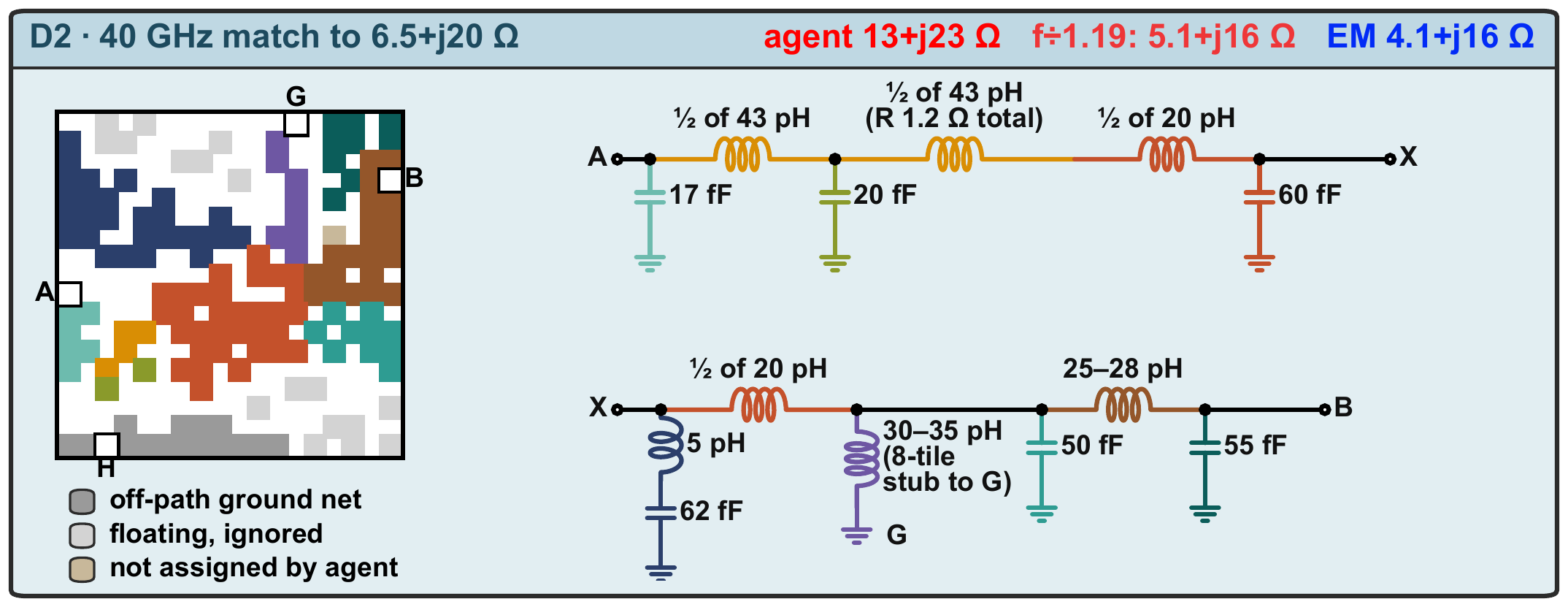}
  \caption{Agent interpretation of GA design D2. Left: layout with tiles colored by the element the agent assigned.
  Right: the agent's stated circuit. Header: $Z_\mathrm{B}$ at the task frequency from the agent's table, from its stated circuit evaluated at $f/1.19$, and from EM.}
  \label{fig:reader_D2}
\end{figure*}

\begin{figure*}[!p]
  \centering
  \includegraphics[width=\textwidth]{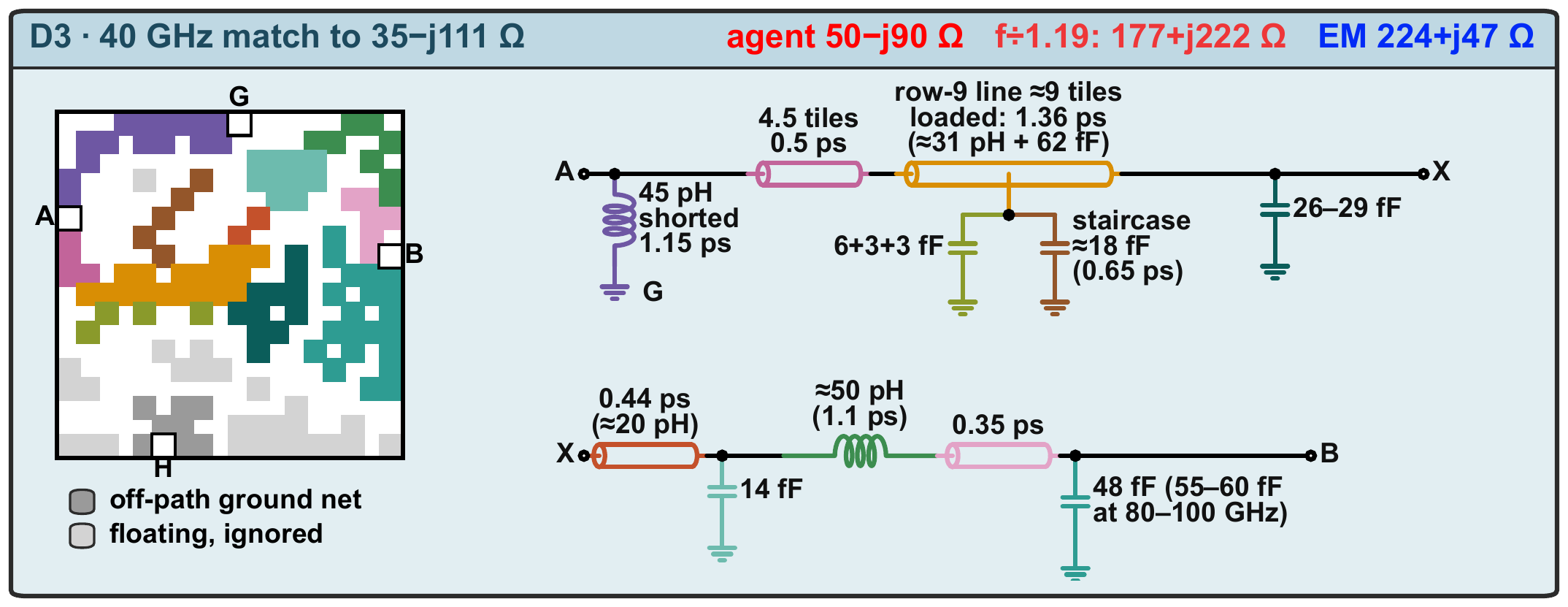}
  \caption{Agent interpretation of GA design D3. Left: layout with tiles colored by the element the agent assigned.
  Right: the agent's stated circuit. Header: $Z_\mathrm{B}$ at the task frequency from the agent's table, from its stated circuit evaluated at $f/1.19$, and from EM.}
  \label{fig:reader_D3}
\end{figure*}

\begin{figure*}[!p]
  \centering
  \includegraphics[width=\textwidth]{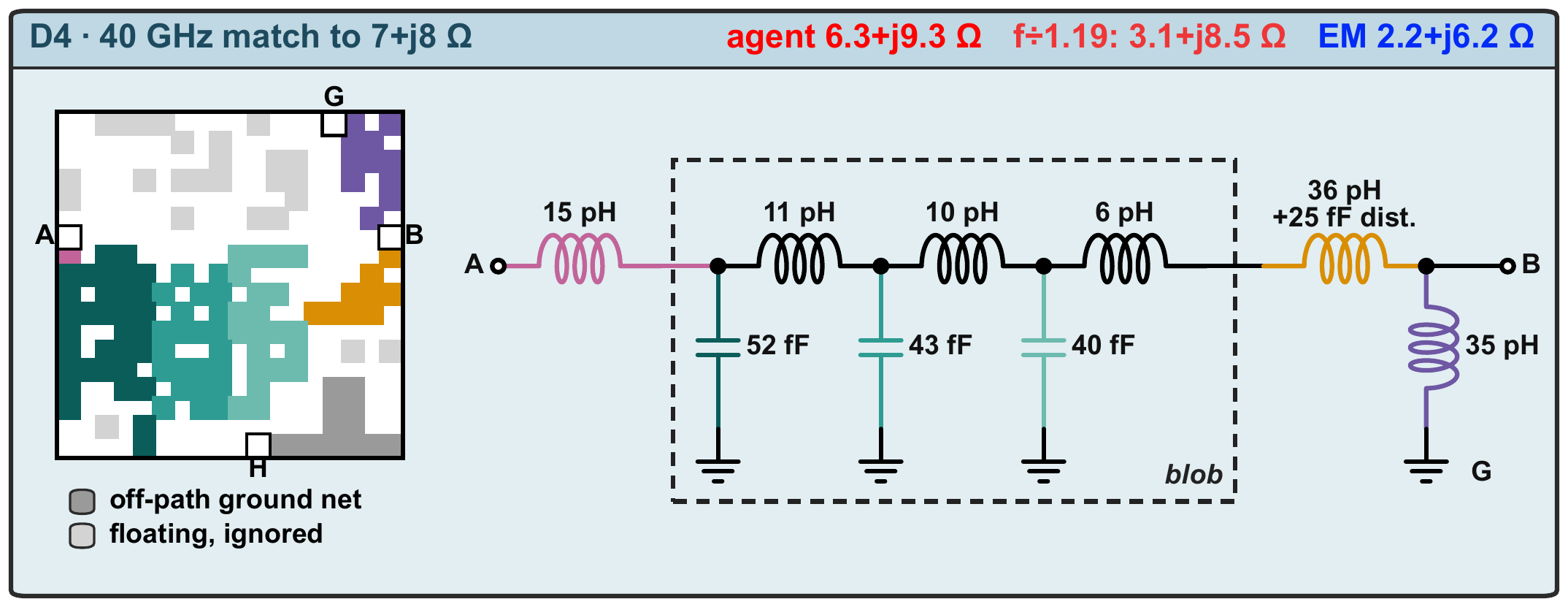}
  \caption{Agent interpretation of GA design D4. Left: layout with tiles colored by the element the agent assigned.
  Right: the agent's stated circuit. Header: $Z_\mathrm{B}$ at the task frequency from the agent's table, from its stated circuit evaluated at $f/1.19$, and from EM.}
  \label{fig:reader_D4}
\end{figure*}

\begin{figure*}[!p]
  \centering
  \includegraphics[width=\textwidth]{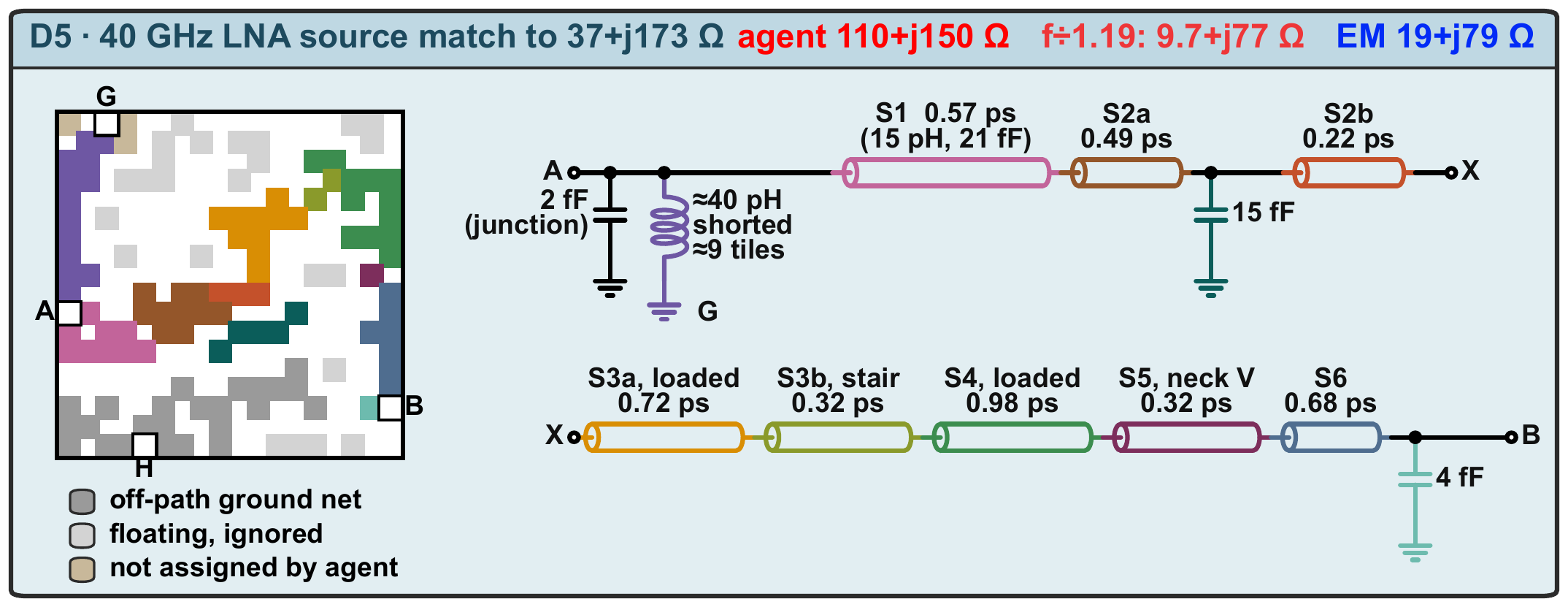}
  \caption{Agent interpretation of GA design D5. Left: layout with tiles colored by the element the agent assigned.
  Right: the agent's stated circuit. Header: $Z_\mathrm{B}$ at the task frequency from the agent's table, from its stated circuit evaluated at $f/1.19$, and from EM.}
  \label{fig:reader_D5}
\end{figure*}

\begin{figure*}[!p]
  \centering
  \includegraphics[width=\textwidth]{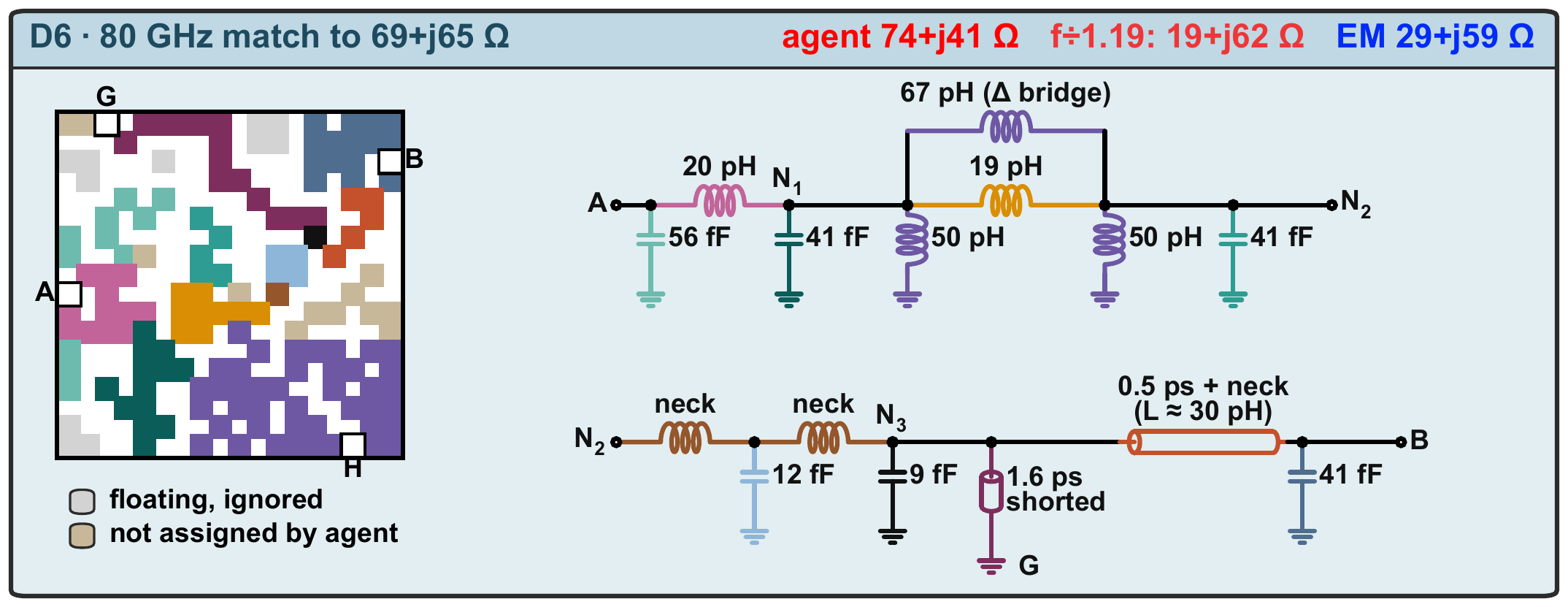}
  \caption{Agent interpretation of GA design D6. Left: layout with tiles colored by the element the agent assigned.
  Right: the agent's stated circuit. Header: $Z_\mathrm{B}$ at the task frequency from the agent's table, from its stated circuit evaluated at $f/1.19$, and from EM.}
  \label{fig:reader_D6}
\end{figure*}

\begin{figure*}[!p]
  \centering
  \includegraphics[width=\textwidth]{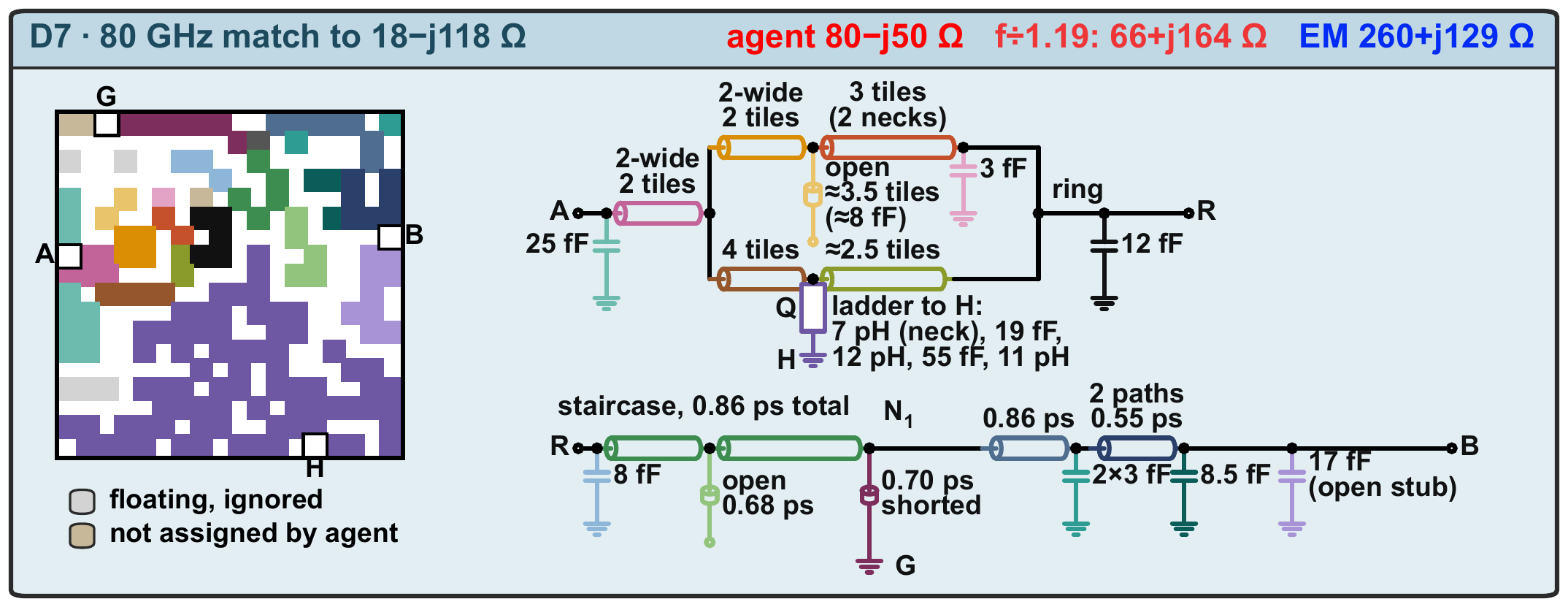}
  \caption{Agent interpretation of GA design D7. Left: layout with tiles colored by the element the agent assigned.
  Right: the agent's stated circuit. Header: $Z_\mathrm{B}$ at the task frequency from the agent's table, from its stated circuit evaluated at $f/1.19$, and from EM.}
  \label{fig:reader_D7}
\end{figure*}

\begin{figure*}[!p]
  \centering
  \includegraphics[width=\textwidth]{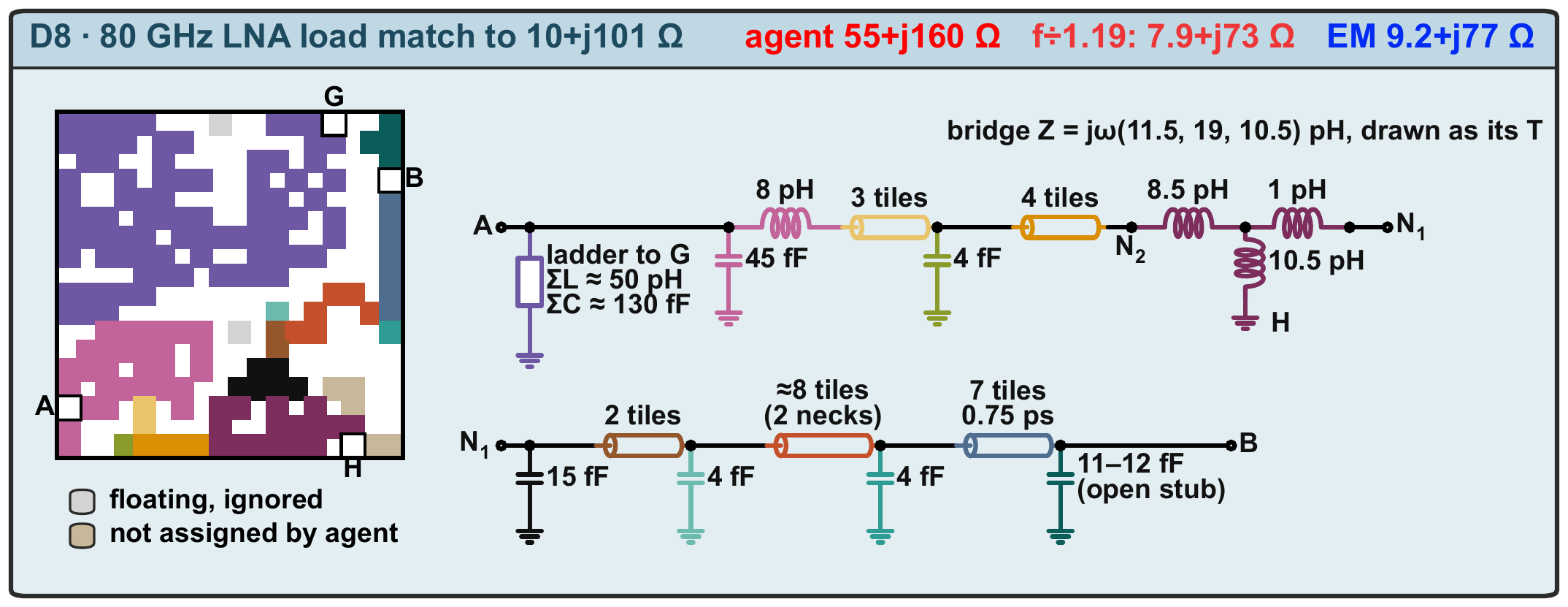}
  \caption{Agent interpretation of GA design D8. Left: layout with tiles colored by the element the agent assigned.
  Right: the agent's stated circuit. Header: $Z_\mathrm{B}$ at the task frequency from the agent's table, from its stated circuit evaluated at $f/1.19$, and from EM.}
  \label{fig:reader_D8}
\end{figure*}

\begin{figure*}[!p]
  \centering
  \includegraphics[width=\textwidth]{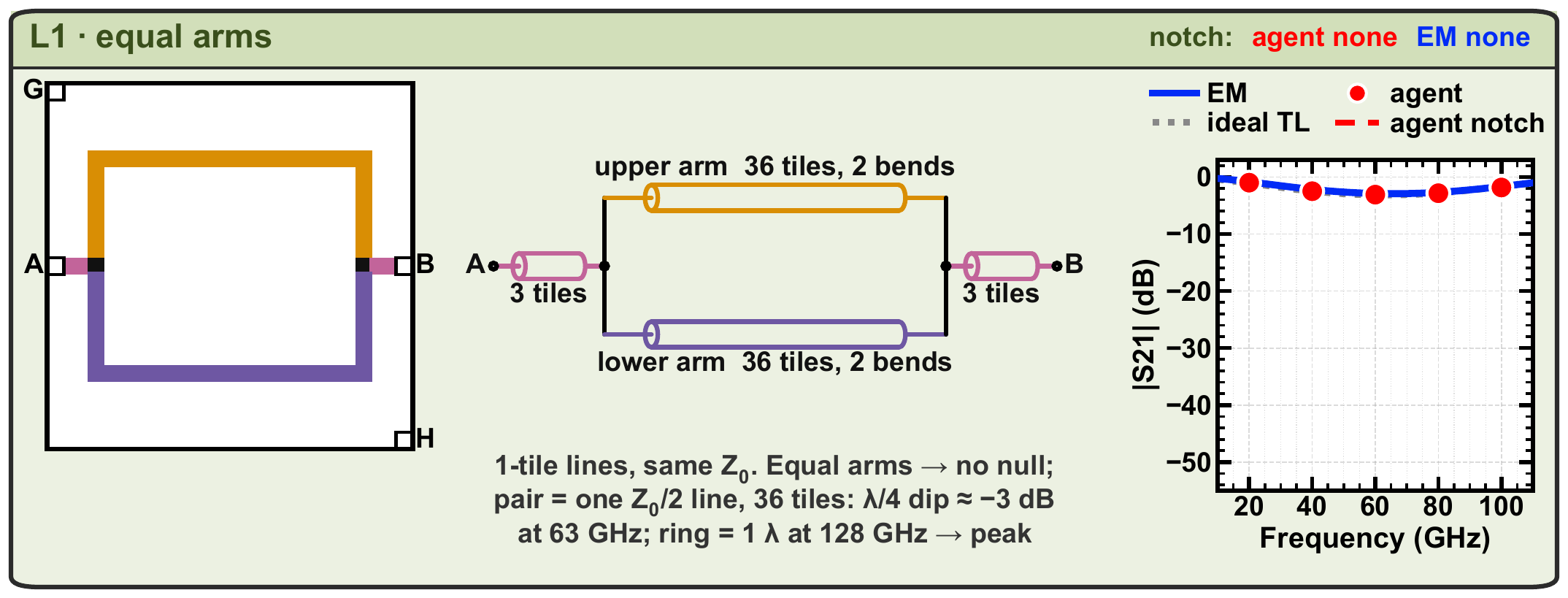}
  \caption{Agent interpretation of loop structure L1: tiles colored by assigned element, the agent's stated
  network, and $|S_{21}|$ from EM, ideal transmission-line theory and the agent.}
  \label{fig:reader_L1}
\end{figure*}

\begin{figure*}[!p]
  \centering
  \includegraphics[width=\textwidth]{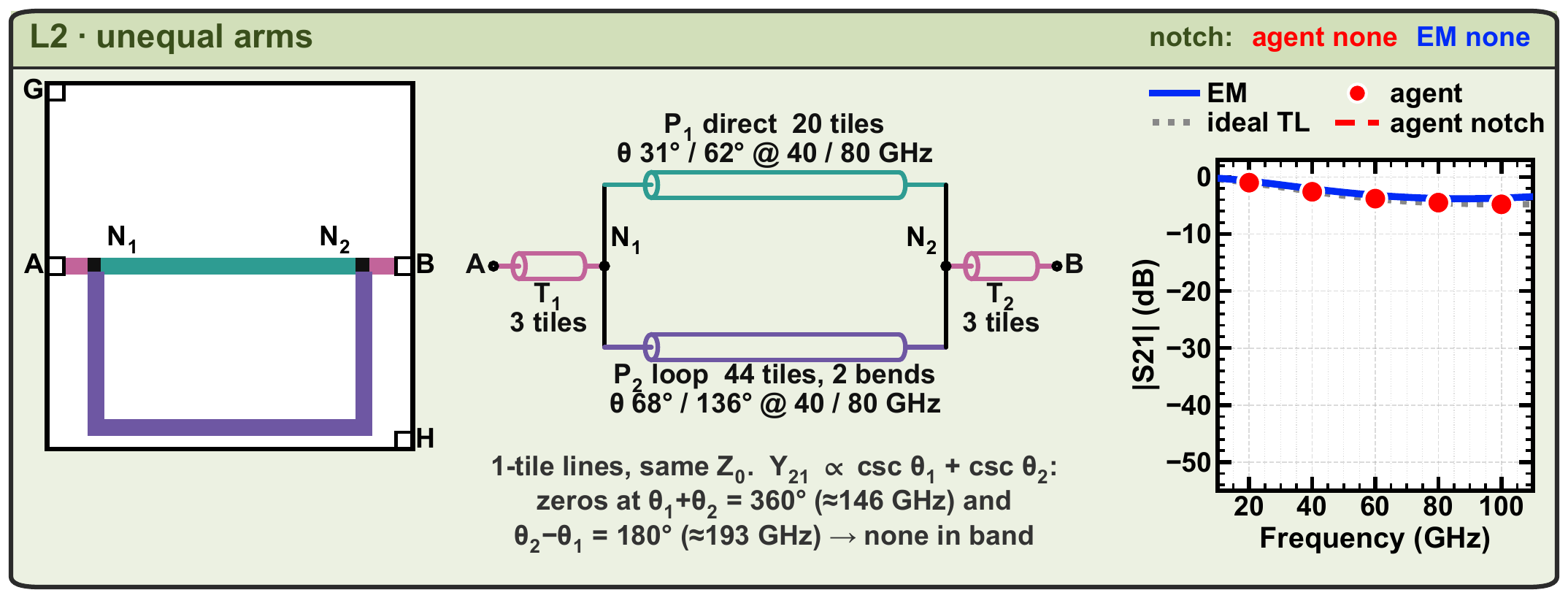}
  \caption{Agent interpretation of loop structure L2: tiles colored by assigned element, the agent's stated
  network, and $|S_{21}|$ from EM, ideal transmission-line theory and the agent.}
  \label{fig:reader_L2}
\end{figure*}

\begin{figure*}[!p]
  \centering
  \includegraphics[width=\textwidth]{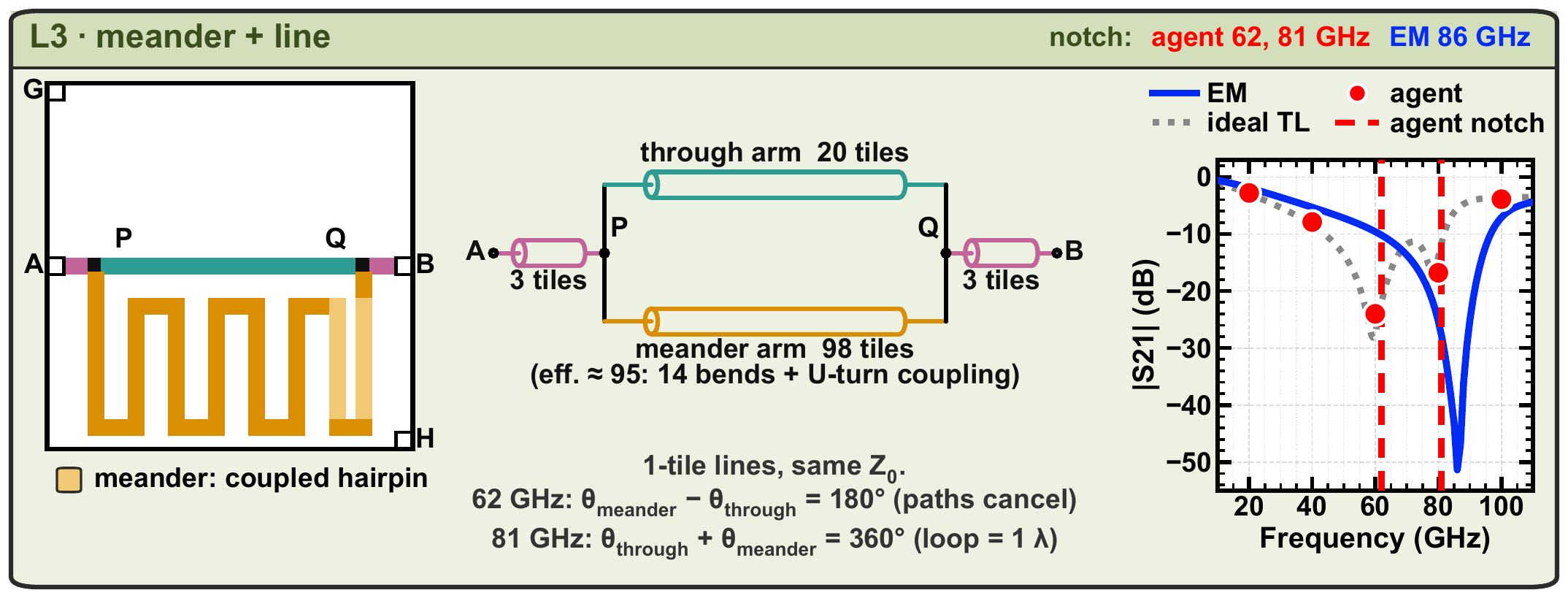}
  \caption{Agent interpretation of loop structure L3: tiles colored by assigned element, the agent's stated
  network, and $|S_{21}|$ from EM, ideal transmission-line theory and the agent.}
  \label{fig:reader_L3}
\end{figure*}

\begin{figure*}[!p]
  \centering
  \includegraphics[width=\textwidth]{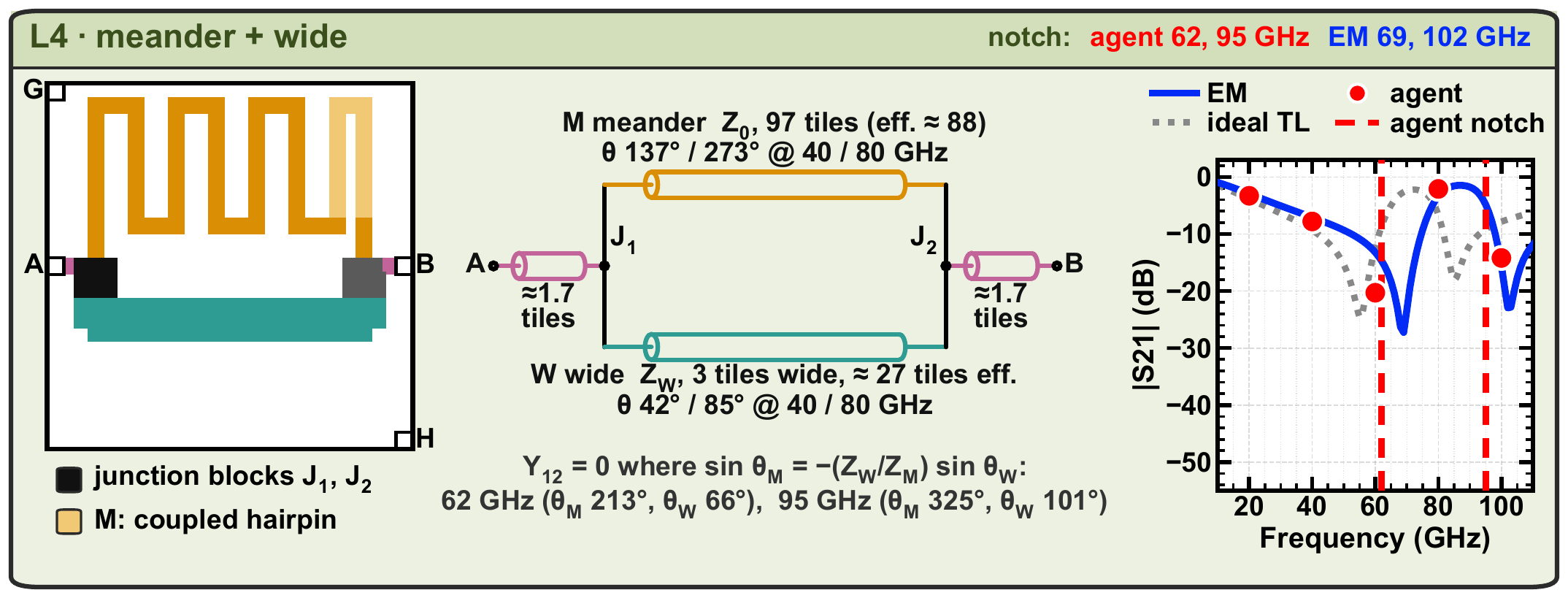}
  \caption{Agent interpretation of loop structure L4: tiles colored by assigned element, the agent's stated
  network, and $|S_{21}|$ from EM, ideal transmission-line theory and the agent.}
  \label{fig:reader_L4}
\end{figure*}

\begin{figure*}[!p]
  \centering
  \includegraphics[width=\textwidth]{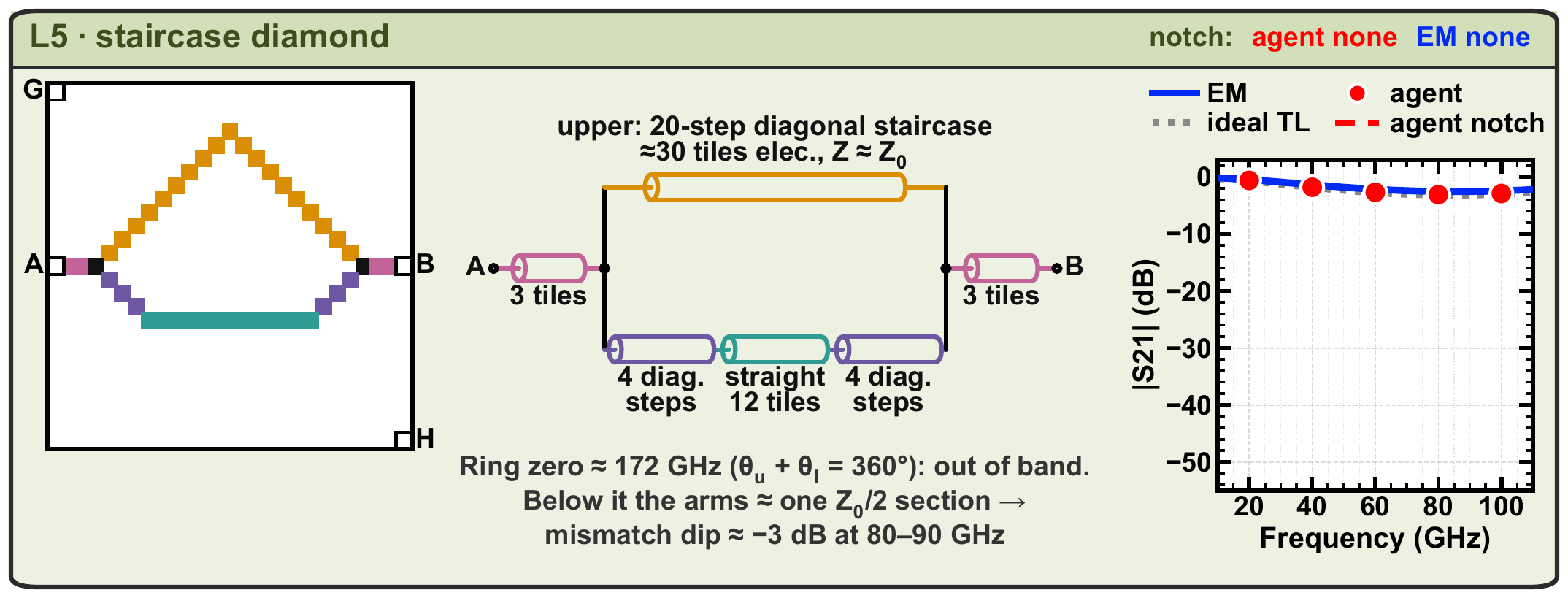}
  \caption{Agent interpretation of loop structure L5: tiles colored by assigned element, the agent's stated
  network, and $|S_{21}|$ from EM, ideal transmission-line theory and the agent.}
  \label{fig:reader_L5}
\end{figure*}

\begin{figure*}[!p]
  \centering
  \includegraphics[width=\textwidth]{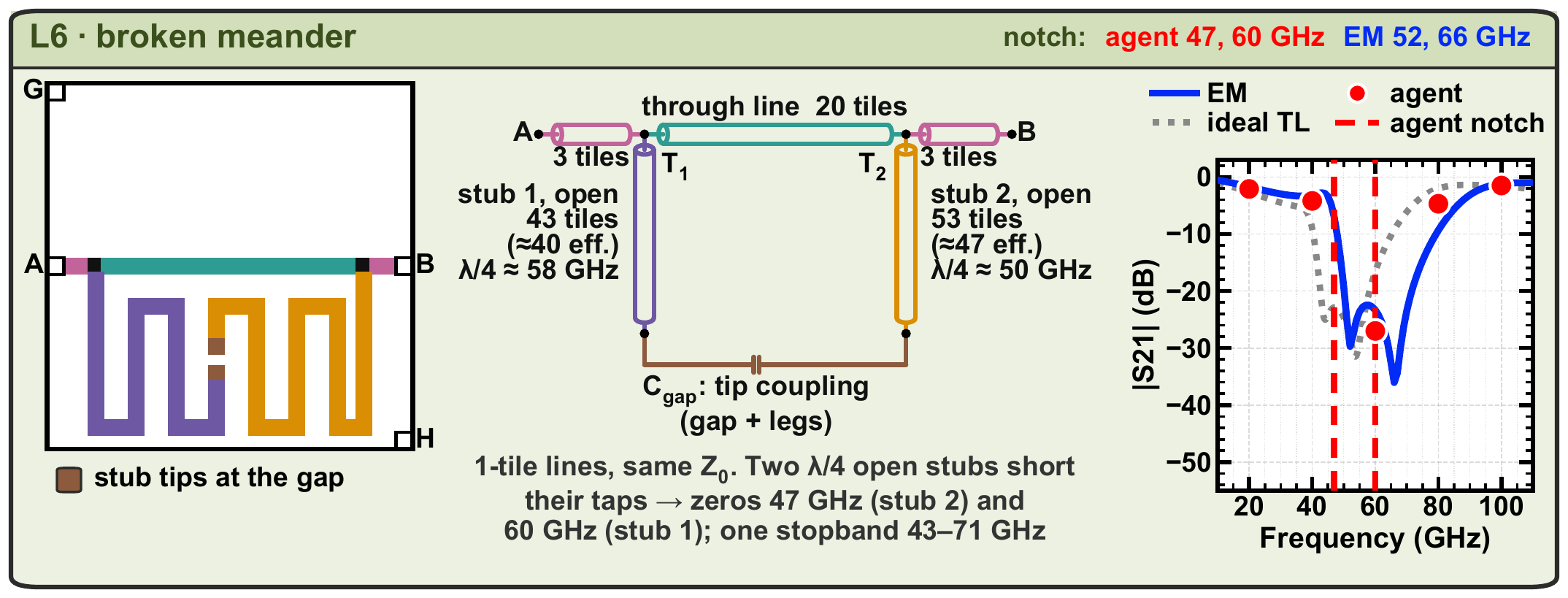}
  \caption{Agent interpretation of loop structure L6: tiles colored by assigned element, the agent's stated
  network, and $|S_{21}|$ from EM, ideal transmission-line theory and the agent.}
  \label{fig:reader_L6}
\end{figure*}

\begin{figure*}[!p]
  \centering
  \includegraphics[width=\textwidth]{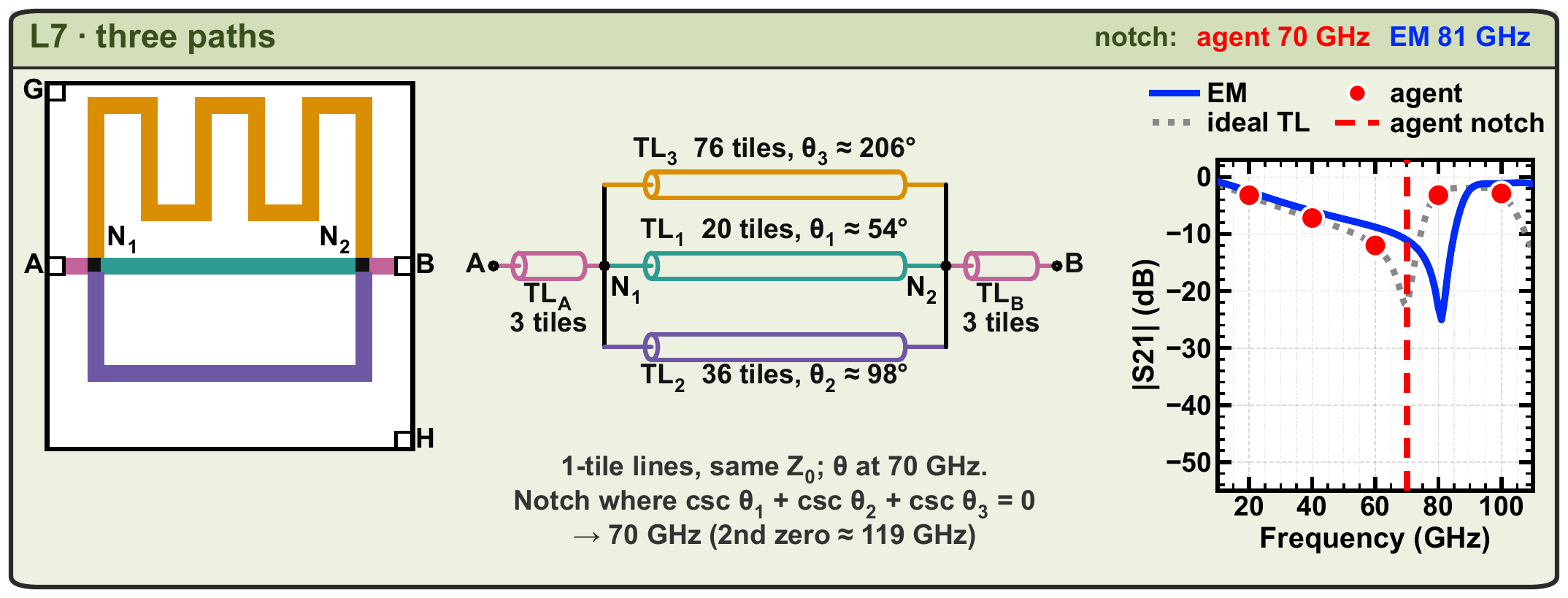}
  \caption{Agent interpretation of loop structure L7: tiles colored by assigned element, the agent's stated
  network, and $|S_{21}|$ from EM, ideal transmission-line theory and the agent.}
  \label{fig:reader_L7}
\end{figure*}

\begin{figure*}[!p]
  \centering
  \includegraphics[width=\textwidth]{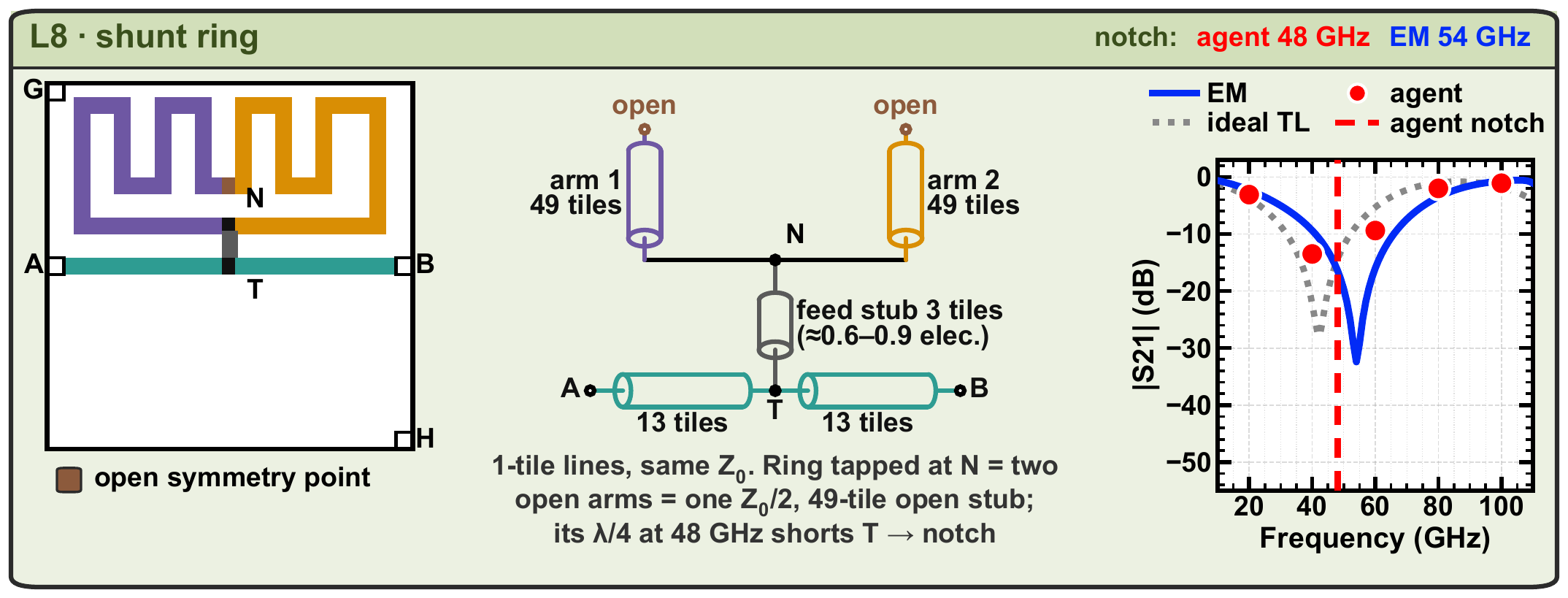}
  \caption{Agent interpretation of loop structure L8: tiles colored by assigned element, the agent's stated
  network, and $|S_{21}|$ from EM, ideal transmission-line theory and the agent.}
  \label{fig:reader_L8}
\end{figure*}

Fig.~\ref{fig:S1} shows the impedance predictions for every read design, Fig.~\ref{fig:S2} all eight loop structures, and
Fig.~\ref{fig:S3} the grid primitives given to the agent as its reference table. Figs.~\ref{fig:reader_D1}--\ref{fig:reader_L8}
show how the agent turned each layout into a circuit: tiles are colored by the element the agent assigned them to (light
grey: called floating; mid grey: ground net), and the circuit is drawn as stated in the agent's answer, with its values. Because the agent's predicted features sit about 16\% low in frequency (Fig.~\ref{fig:prediction}(e)), a single-frequency impedance can land far around the Smith chart from EM even when the contour shape is right. The headers of Figs.~\ref{fig:reader_D1}--\ref{fig:reader_D8} therefore also give the agent's stated circuit at $f/1.19$, the GA group median $s^*$, fitted post hoc on the same designs: across D1--D8 this lowers the mean $|\Delta\Gamma|$ at the task frequency from 0.45 to 0.14 and removes all three reactance-sign mismatches.

\end{document}